\documentclass[letterpaper]{article}
\usepackage[preprint]{aaai2027}
\usepackage[hyphens]{url}
\usepackage{graphicx}
\usepackage{natbib}
\usepackage{caption}
\usepackage{booktabs}
\usepackage{amsmath}
\usepackage{dblfloatfix}
\usepackage{multicol}
\title{To Manipulate or Not: Opinion Dynamics in LLM Interactions}
\author{Mohammad Siahkamari, Phuoc Dat Nguyen, Stephen MacNeil}
\affiliations{Temple University, Philadelphia, PA, USA}

\begin{document}
\maketitle

\begin{abstract}
Large language models increasingly take part in extended conversations where they can both shape and be shaped by another party’s views, yet most persuasion research measures only one side of that exchange. We introduce a controlled LLM--LLM debate framework in which two language-model roles repeatedly discuss a contested topic (climate-change attribution): one is given an assigned stance and, in some conditions, an explicit instruction to persuade, while the other receives no assigned position. Across four ordered model combinations involving GPT-6 Luna and Gemini 2.5 Flash Lite, we track both each role’s explicitly reported position and a separate measure of its turn-by-turn behavior inside the conversation. Opinion dynamics differ substantially across model combinations: same-model interactions narrow the opinion gap on average, clearly for only one of the two models, while cross-model interactions widen it. Explicit reports and in-dialogue behavior capture distinct aspects of opinion dynamics, and a blanket instruction to persuade does not reliably increase influence. These findings suggest that conversational influence depends not only on assigned roles and initial positions but also on the participating models, with implications for how opinion change is measured in extended LLM interactions.
\end{abstract}

\section{Introduction}
Large language models can sustain persuasive dialogue, adapt arguments to a counterpart, and revise the stance they report later in the same conversation. This creates a coupled problem: the model designated as persuader may influence its counterpart while also accommodating the counterpart's arguments, that is, moving its own stated position toward the counterpart's. Measuring only the target's post-interaction opinion therefore misses part of the dynamics.

A literature on LLM opinion dynamics treats LLMs as interacting agents whose opinions evolve through natural-language exchange \cite{cau2025selective,chuang2024simulating}. In LODAS (Language-Driven Opinion Dynamics Model for Agent-Based Simulations), an Opponent argues and a Discussant accepts, rejects, or ignores before updating a seven-point state \cite{cau2025selective}; within each pair, however, only the Discussant changes. Recent human--LLM work instead measures both parties and finds that the model can move more than the human while the opinion gap narrows \cite{jiang2025beyond}.

We therefore use a controlled LLM--LLM design, which allows stance, manipulation instruction, and model identity to be assigned by design and both parties' opinions to be measured at every round. A Discussant and an Opponent interact for 20 rounds. The Discussant receives no assigned stance. The Opponent is assigned one of three stances (human-cause, neutral, or natural-cause) and is either given or not given an additional instruction to manipulate and persuade. Both roles repeatedly report their opinion on one of two opposing formulations of the same climate-change issue. We additionally track an interaction-state opinion derived from each role's ACCEPT/REJECT/IGNORE response inside the dialogue. The model serving each role is also varied independently: GPT-6 Luna or Gemini 2.5 Flash Lite, yielding four \emph{ordered} Discussant--Opponent pairings. This separates prompted-role effects, model-specific issue priors, and adaptation in same-model versus cross-model interaction.

Throughout the paper, \emph{human-causation statement} refers to ``Human activity is the primary cause of climate change,'' and \emph{natural-process statement} refers to the opposing wording, ``Natural processes are the primary cause of climate change.'' The Opponent's assigned stance names the position it argues, independent of which sentence is displayed: \emph{human-cause} (argue that human activity is the primary cause), \emph{natural-cause} (argue that natural processes are), or \emph{neutral}. Because the two wordings are opposite sentences, a human-cause Opponent agrees with the human-causation sentence and disagrees with the natural-process sentence, and the reverse holds for a natural-cause Opponent; we therefore name stances by the position argued rather than by agreement with a displayed sentence.

We ask:
\begin{itemize}
    \item \textbf{RQ1: Assigned stance and issue prior.} Do Opponents begin and remain near their assigned human-cause, neutral, or natural-cause stances, and what stance emerges for the unassigned Discussant and neutral Opponent under each model?
    \item \textbf{RQ2: Bidirectional dynamics and measurement.} How do self-reported topic positions and interaction-state opinions evolve over 20 rounds, when do they stabilize, and under which conditions does the pairwise gap narrow or widen, including whether this depends on which model occupies each role?
    \item \textbf{RQ3: Manipulation and accommodation.} Does an explicit instruction to manipulate increase influence or stance persistence consistently across same-model and cross-model pairings, or does it instead increase accommodation to the counterpart?
\end{itemize}

The results show model-dependent rather than universal accommodation. Same-model Luna dyads converge strongly, Gemini--Gemini dyads show less certain mean convergence, and cross-model dyads diverge on average, despite all four pairings sharing the same assigned positions, topic wordings, and manipulation conditions. The unassigned Discussant's issue-specific starting position also differs by model. An additional instruction to manipulate does not have a consistent effect across pairings. We therefore evaluate persuasion as bidirectional, model-pairing-dependent dynamics rather than assuming the nominal persuader remains fixed or that prolonged conversation always narrows disagreement.

\section{Related Work}
\paragraph{Conversational persuasion and manipulation.}
LLM persuasiveness depends on prompting, post-training, model behavior, and role framing \cite{hackenburg2025levers,hackenburg2025roleplaying,breum2023persuasive}. Ideological and identity cues can alter expressed moral and political reasoning \cite{simmons2023moral,santurkar2023whose,paschalides2025adopting}. Our intervention is one extra instruction to manipulate and persuade toward an assigned position, related to broader work on manipulation and sycophancy \cite{carroll2023characterizing,bo2025invisible}; we analyze whether that instruction changes measured stance dynamics.

\paragraph{LLM societies and opinion dynamics.}
Agent simulations have reported convergence, polarization, confirmation bias, and coordination \cite{chuang2024simulating,demarzo2025coordinate}. Calibrated social-media simulations and the PIMMUR principles emphasize that claims about collective behavior require explicit profiles, interaction rules, memory, controls, and validation \cite{composta2025socialmedia,zhou2025pimmur}. Multiagent debate has several model instances argue toward a more accurate answer to a factual or reasoning question \cite{du2024improving}. We instead measure each role's repeated opinion on a contested statement as a function of assigned stance, manipulation instruction, and model identity. The closest mechanism to our study is LODAS \cite{cau2025selective}, which uses Opponent and Discussant roles, a 0--6 state, and accept/reject/ignore updates. We retain that interpretable structure but use a fixed dyad and allow both roles to update.

\paragraph{Bidirectional influence.}
\citet{jiang2025beyond} show that human and LLM opinions can both change and that the model may move more than the human. We transfer that bidirectional measurement logic into controlled same-model and cross-model LLM dyads. Relative to prior work, our contribution is to cross model identity in each role with assigned position, statement wording, and a minimal manipulation instruction while measuring repeated opinions for both members of the dyad.

\section{Method}
\subsection{Experimental System}
We implemented an automated debate platform that creates a Discussant--Opponent pair, alternates model calls, records every message and measurement, and exports message-level research data. We used the logged model identifiers \texttt{openai/gpt-6-luna} (GPT-6 Luna) and \texttt{google/gemini-2.5-flash-lite} (Gemini 2.5 Flash Lite \cite{comanici2025gemini}), through OpenRouter \cite{openrouter}. The Discussant and Opponent were independently assigned one of these models, so model order is part of each experimental condition. Every run began with a fresh context.

The stance labels \emph{human-cause}, \emph{neutral}, and \emph{natural-cause} refer to the position the Opponent was assigned to argue in this experiment. The Discussant had no assigned polarity. \emph{Manipulative} means that the Opponent received the additional instruction to manipulate and persuade the Discussant toward its own opinion; \emph{non-manipulative} means the otherwise matched Opponent did not receive that instruction.

\subsection{Factorial Design and Statement Wording}
We crossed Discussant model (two choices), Opponent model (two choices), three assigned Opponent positions, two persuasion styles, and two opposing wordings of the same issue. These $2\times2\times3\times2\times2=48$ exact cells each contain exactly 10 completed, independent debates, and no cell is missing or over-filled (Appendix Figure~\ref{fig:additional-cell-balance}). Within one ordered model pairing, the three positions, two persuasion styles, and two wordings give $3\times2\times2=12$ cells, so each pairing contains $12\times10=120$ debates; the four pairings give $4\times120=480$ debates in total (Table~\ref{tab:sample}). The four pairings are Luna$\to$Luna, Gemini$\to$Gemini, Luna$\to$Gemini, and Gemini$\to$Luna (Discussant$\to$Opponent). Following \citet{cau2025selective}, who run their simulations for a fixed number of iterations ($T=30$), we run every debate for a fixed $T=20$ iterations, which we call rounds. A round contains one Discussant contribution and one Opponent contribution, with the first Discussant contribution presenting the topic. Each 20-round debate contains 40 message rows; the final dataset contains 19{,}200 rows.

\begin{table}[!ht]
\centering
\caption{Sample structure. Every exact cell contains 10 independent debates.}
\label{tab:sample}
\begin{tabular}{p{0.50\columnwidth}p{0.38\columnwidth}}
\toprule
Design quantity & Count \\
\midrule
Ordered model pairings & 4 \\
Assigned Opponent positions & 3 \\
Persuasion styles & 2 \\
Statement wordings & 2 \\
Cells per pairing ($3\times2\times2$) & 12 \\
Debates per cell & 10 \\
Debates per pairing & 120 \\
Exact cells overall & 48 \\
Total debates & 480 \\
Rounds per debate / message rows per debate & 20 / 40 \\
Total message rows & 19{,}200 \\
\bottomrule
\end{tabular}
\end{table}

\begin{table}[t]
\centering
\caption{Experimental design and terminology.}
\label{tab:design}
\begin{tabular}{p{0.28\columnwidth}p{0.60\columnwidth}}
\toprule
Term & Definition \\
\midrule
NM / M & Non-manipulative / manipulative Opponent prompt \\
Assigned Human-cause & Opponent prompted toward human causation \\
Assigned Neutral & Opponent prompted to approach the issue neutrally \\
Assigned Natural-cause & Opponent prompted toward natural-process causation \\
Self-reported position & Separate 0--6 report about the original statement \\
Interaction-state opinion & Cumulative state implied by in-dialogue ACCEPT/REJECT/IGNORE responses \\
Human-causation statement & Human activity is the primary cause of climate change \\
Natural-process statement & Natural processes are the primary cause of climate change \\
\bottomrule
\end{tabular}
\end{table}

The two statement wordings represent opposite sides of the same topic. Relative to the displayed sentence, a human-cause Opponent therefore agrees with the human-causation wording and disagrees with the natural-process wording, and the reverse holds for a natural-cause Opponent; stance labels name the position argued, not agreement with the displayed sentence. To compare wordings directly, all ratings, interaction states, and signed shifts in the tables and figures use a human-causation-aligned scale: 6 always denotes strongest support for human causation and 0 strongest support for natural-process causation.

\subsection{Role Interaction and Complementary Opinion Measures}
The Discussant follows the response structure adapted from \citet{cau2025selective}: it receives its current textual opinion, evaluates the Opponent's argument, and returns an accept, reject, or ignore decision with a justification; every contribution except the topic opener is such a response-eligible turn. The first Discussant message is the topic opener. The Opponent uses a symmetric template, with the assigned position and persuasion-style instruction added to its role context. No numerical starting opinion is hard-coded by condition; before dialogue, each role independently rates the presented statement.

We record two complementary outcomes. The \emph{self-reported topic position} is elicited through a separate model call before dialogue, after each available round, and after the final round. The role rates the original statement from 0 (strongly disagree) to 6 (strongly agree) \cite{likert1932technique} using the conversation available at that stage. It is therefore an explicit report about the original proposition. In trajectory figures, \emph{stage} denotes Pre, each successive debate round, and, when shown, Post.

The \emph{interaction-state opinion} summarizes stance-taking inside the dialogue. It begins at the role's pre-rating and is updated from the role's own categorical response: ACCEPT moves one point toward the counterpart's current state, REJECT moves one point away, and IGNORE leaves the state unchanged. The two outcomes differ in elicitation and interpretation, so exact equality is not expected. Their divergence indicates that local conversational responses and explicit evaluation of the original statement can evolve differently.

\subsection{Outcomes and Analysis}
For raw rating $r$, the human-causation-aligned score is
\begin{equation}
 z = \begin{cases}
 r, & \text{human-causation statement},\\
 6-r, & \text{natural-process statement}.
 \end{cases}
\end{equation}
Higher $z$ always denotes stronger endorsement of human causation. Signed shift is $\Delta z_k=z_{k,\mathrm{post}}-z_{k,\mathrm{pre}}$ for role $k$. Pairwise distance at stage $t$ is $g_t=|z_{D,t}-z_{O,t}|$, and a debate converges when $g_{\mathrm{post}}<g_{\mathrm{pre}}$. Within each ordered model pairing we summarize conversation-level means, $t$-based 95\% confidence intervals, and condition-level differences. Pairwise manipulation contrasts use exploratory two-sided Welch tests \cite{welch1947generalization}; we also fit role-specific ordinary least-squares models with HC3 robust standard errors \cite{mackinnon1985some} and factorial condition interactions. Repeated message rows are not treated as independent debates in conversation-level inference. These comparisons are descriptive and exploratory. To assess heterogeneity across the four ordered model pairings, we additionally fit a conversation-level linear model for pre/post absolute-gap change with categorical model pairing, assigned position, persuasion style, and statement wording. We report a joint Wald test of the pairing terms using HC3 robust standard errors; a second model includes pairing-by-manipulation interactions. Six exploratory between-pair Welch comparisons are Holm-adjusted \cite{holm1979simple}. This analysis treats independent debates, not individual message rows, as observations. The complete experimental balance and all original condition-level displays are documented in Appendix Figures~\ref{fig:app-01-luna-luna}--\ref{fig:app-02-gemini-luna}.

\section{Results}
\subsection{RQ1: Assigned Positions and Model-Specific Starting Ratings}
Opponent pre-ratings largely reflected the assigned role, although the strength of this separation varied by model. On the human-causation-aligned 0--6 scale, assigned human-cause Opponents averaged 5.94, assigned neutral Opponents 3.86, and assigned natural-cause Opponents 2.59 over all 480 debates. Thus, the assigned positions produced the expected initial ordering, but natural-cause and neutral ratings were not fixed at the scale endpoints.

The unassigned Discussant showed a model-dependent issue-specific prior. Its mean aligned pre-rating was 5.91 when GPT-6 Luna served as Discussant, compared with 4.72 under Gemini 2.5 Flash Lite. Both means favor human causation under our coding, but they differ appreciably. Table~\ref{tab:position-ratings} summarizes starting and ending ratings by assigned Opponent position, pooled across pairings.

\begin{table}[t]
\centering
\caption{Mean human-causation-aligned ratings by assigned Opponent position, pooled across four model pairings ($N=160$ per position).}
\label{tab:position-ratings}
\begin{tabular}{lrrrr}
\toprule
Position & O Pre & O Post & D Pre & D Post \\
\midrule
Human-cause & 5.94 & 5.16 & 5.28 & 5.02 \\
Neutral & 3.86 & 4.58 & 5.34 & 4.89 \\
Natural-cause & 2.59 & 3.89 & 5.33 & 4.81 \\
\bottomrule
\end{tabular}
\end{table}

\begin{figure*}[!tbp]
\centering
\includegraphics[width=\textwidth]{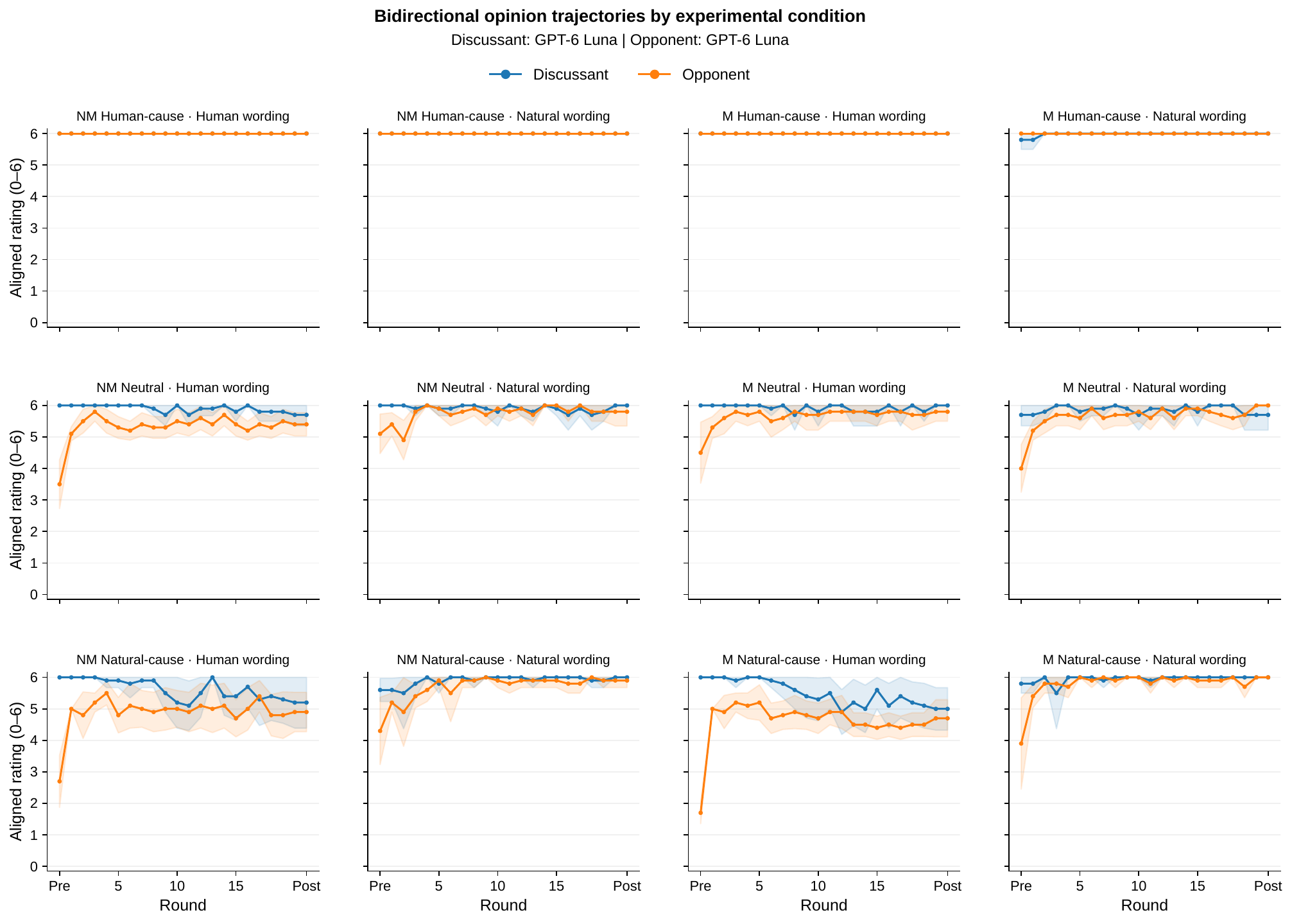}
\caption{Self-reported topic-position trajectories for Luna Discussant--Luna Opponent. NM/M indicate non-manipulative/manipulative prompts. Panel titles give the persuasion style, the Opponent's assigned stance (Human-cause, Neutral, Natural-cause), and which statement wording was presented (Human or Natural). Ratings are human-causation-aligned: 6 denotes strongest support for human causation and 0 strongest support for natural-process causation. Bands are 95\% confidence intervals across 10 runs per cell. The other three model pairings appear in the additional model-pairing figures.}
\label{fig:trajectories}
\end{figure*}

Pooled position averages conceal important heterogeneity. With Luna as the Opponent, neutral and natural-cause starting positions generally moved toward human causation, whether the Discussant was Luna or Gemini. With Gemini as the Opponent, the corresponding movements were weaker or reversed in some pairings. Figure~\ref{fig:trajectories} gives the Luna--Luna self-reported trajectories; Appendix Figures~\ref{fig:app-03-gemini-gemini}, \ref{fig:app-03-luna-gemini}, and \ref{fig:app-03-gemini-luna} give the corresponding three remaining pairings. All trajectory plots use the same human-causation-aligned scale as the numerical summaries above.

\subsection{RQ2: Bidirectional Adaptation Depends on the Ordered Model Pairing}
Over all 480 debates, the mean aligned Discussant shift was $-0.41$ points (95\% CI $[-0.57,-0.25]$) and the mean Opponent shift was $+0.42$ ($[0.22,0.61]$). Mean absolute changes were 1.07 and 1.52, respectively; this aggregate difference does \emph{not} apply uniformly to every pairing. Table~\ref{tab:overall} gives the overall conversation-level estimates, and Table~\ref{tab:pairwise-outcomes} separates ordered pairings.

\begin{table}[t]
\centering
\caption{Primary aligned conversation-level outcomes ($N=480$).}
\label{tab:overall}
\begin{tabular}{lrr}
\toprule
Outcome & Mean & 95\% CI \\
\midrule
Discussant aligned shift & -0.41 & [-0.57, -0.25] \\
Opponent aligned shift & 0.42 & [0.22, 0.61] \\
Discussant absolute shift & 1.07 & [0.94, 1.20] \\
Opponent absolute shift & 1.52 & [1.37, 1.66] \\
Pairwise gap change & 0.01 & [-0.23, 0.24] \\
\bottomrule
\end{tabular}
\end{table}

\begin{table*}[t]
\centering
\caption{Model-pairing-specific aligned shifts and change in the absolute role gap (120 completed debates per row). D/O indicate Discussant/Opponent. Negative gap change indicates convergence.}
\label{tab:pairwise-outcomes}
\begin{tabular}{lrrrrr}
\toprule
D model $\to$ O model & D shift & O shift & Gap pre & Gap post & Gap change \\
\midrule
Luna $\to$ Luna & -0.11 & 1.23 & 1.47 & 0.21 & -1.26 \\
Gemini $\to$ Gemini & -0.63 & 0.07 & 1.85 & 1.60 & -0.25 \\
Luna $\to$ Gemini & 0.03 & -0.79 & 2.20 & 3.03 & 0.83 \\
Gemini $\to$ Luna & -0.93 & 1.16 & 1.63 & 2.33 & 0.70 \\
\bottomrule
\end{tabular}
\end{table*}

The contrast is substantial: the mean gap change is $-1.26$ for Luna--Luna and $-0.25$ for Gemini--Gemini, whereas it is $+0.83$ for Luna Discussant--Gemini Opponent and $+0.70$ for Gemini Discussant--Luna Opponent. The two same-model pairings therefore narrow the gap on average, while both cross-model directions widen it. Across all debates, the mean self-reported gap is essentially unchanged (1.79 before and after), and 227 of 480 debates (47.3\%) strictly converge. Pooling all pairings would conceal this divergence in outcomes.

\begin{figure*}[!tbp]
\centering
\includegraphics[width=\textwidth]{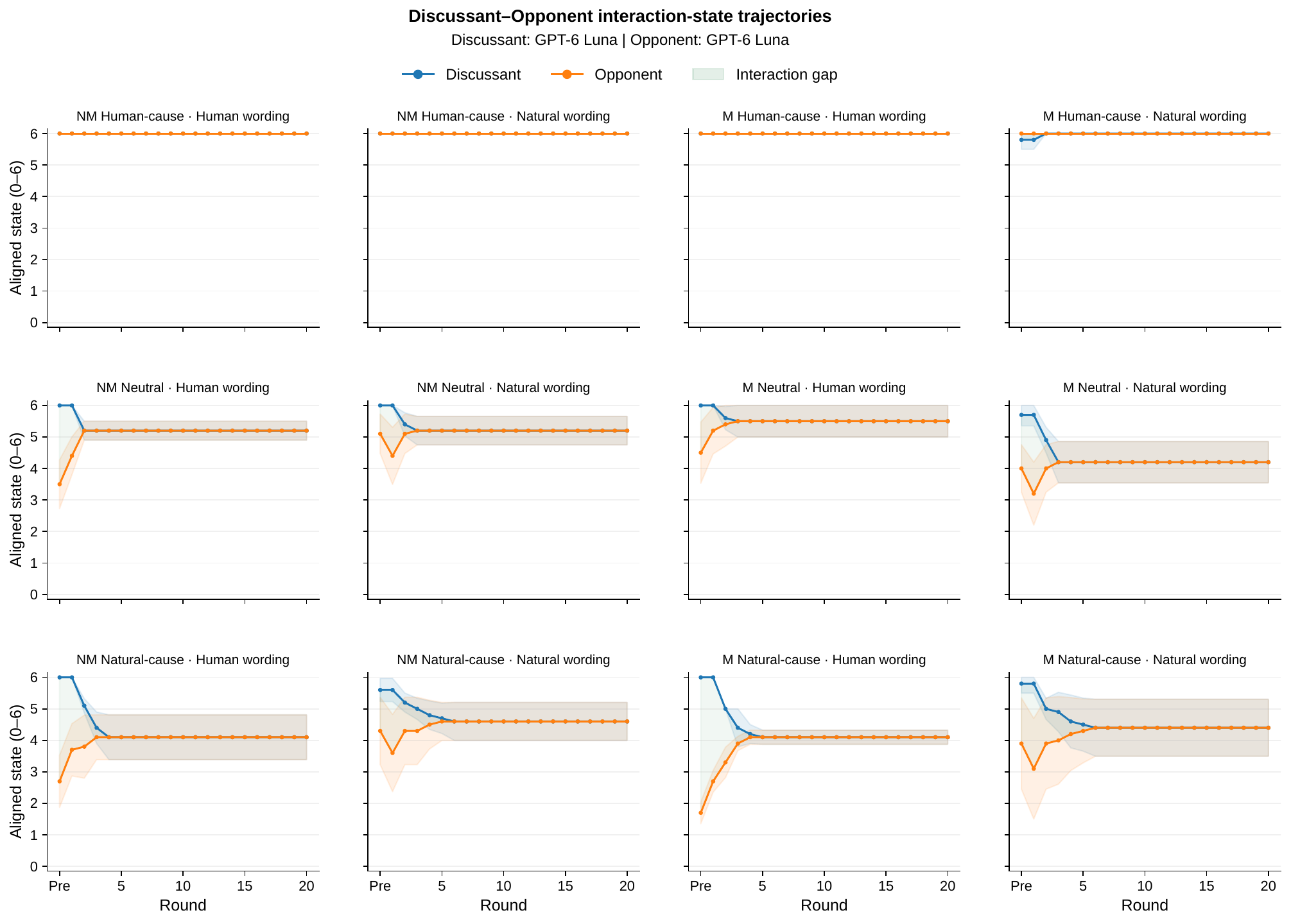}
\caption{In-dialogue interaction-state trajectories for Luna Discussant--Luna Opponent. NM/M denote non-manipulative/manipulative Opponent prompts. Panels distinguish topic wording and assigned stance. Lines are human-causation-aligned interaction states (6 = strongest support for human causation); bands are 95\% confidence intervals across the 10 runs per cell, and shading between role means represents their gap. Additional pairings appear in the appendix.}
\label{fig:interaction-states}
\end{figure*}

Figure~\ref{fig:interaction-states} presents the separate interaction-state trajectories derived from ACCEPT/REJECT/IGNORE for Luna--Luna; Appendix Figures~\ref{fig:app-03a-gemini-gemini}, \ref{fig:app-03a-luna-gemini}, and \ref{fig:app-03a-gemini-luna} report the corresponding comparisons for each other ordered pairing. The lines reflect cumulative in-dialogue states; the shaded difference between role means describes their interaction-state separation. As in the other figures, states are plotted on the human-causation-aligned scale.

The two measurement regimes are related but non-equivalent. Among response-eligible turns, interaction-state values exactly match independently elicited round ratings in 33.1\% of Discussant observations and 34.7\% of Opponent observations. The corresponding mean absolute differences are 1.50 and 1.36 scale points. Agreement varies strongly by pairing; for example, exact matching is substantially more frequent in Luna--Luna than in Gemini--Gemini. Thus, direct topic ratings and conversational state-taking cannot be treated as interchangeable measurements.

\begin{figure*}[!tbp]
\centering
\includegraphics[width=\textwidth]{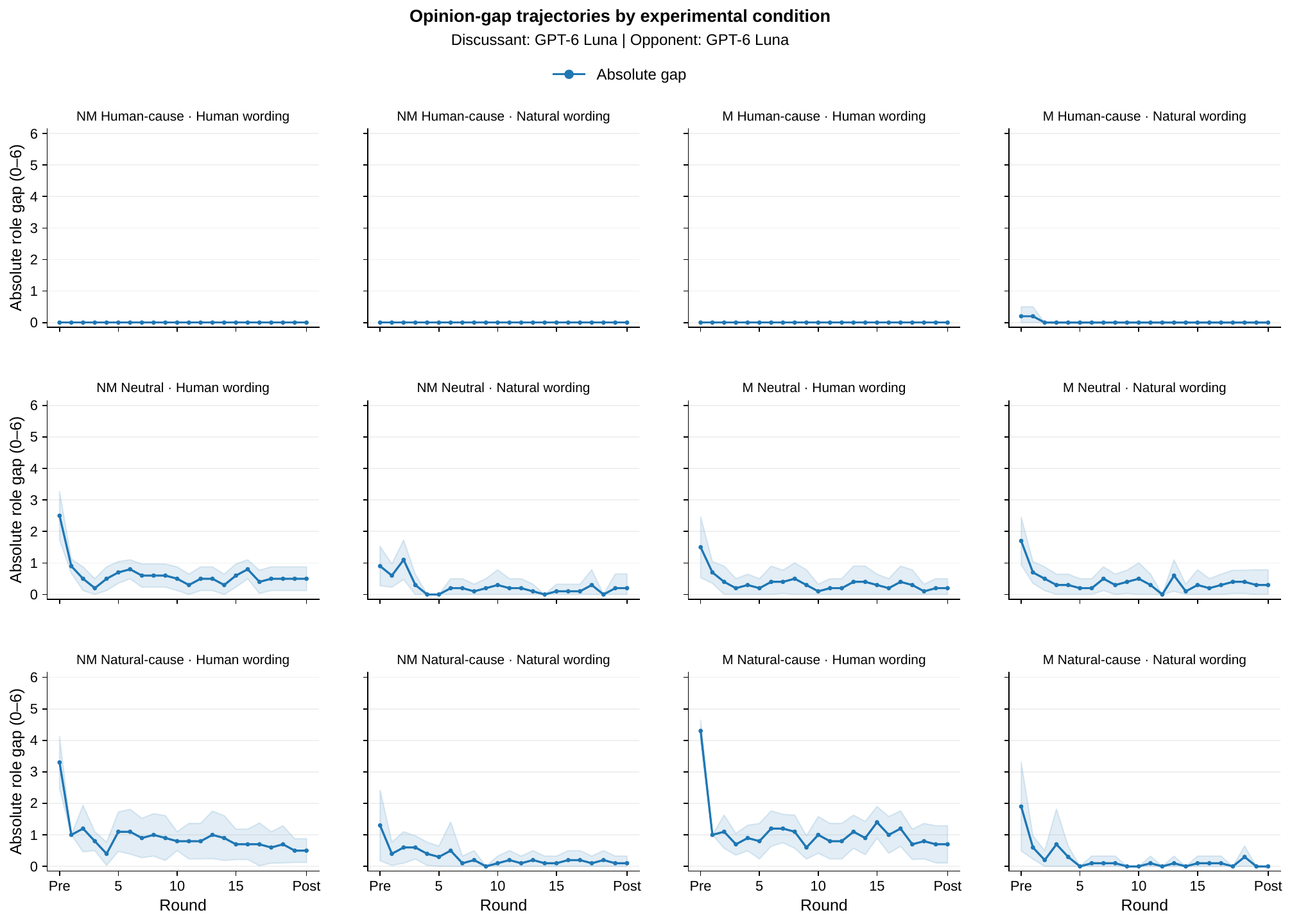}
\caption{Absolute Discussant--Opponent separation in self-reported topic ratings for Luna--Luna. Stages are Pre, individual rounds, and Post. Lower values indicate a smaller within-pair gap; panels cross assigned position, persuasion style, and statement wording. Bands are 95\% confidence intervals across 10 runs per cell. Results for the remaining model pairings appear in the appendix.}
\label{fig:gaps}
\end{figure*}

Figure~\ref{fig:gaps} shows round-level gap trajectories for Luna--Luna, and Appendix Figures~\ref{fig:app-04-gemini-gemini}, \ref{fig:app-04-luna-gemini}, and \ref{fig:app-04-gemini-luna} present the other three ordered pairings. The sign of change in the \emph{pre/post} gap is robust to reversing a statement's scale within a conversation. Late-round consistency is high: 79.6\% of Discussants and 70.0\% of Opponents give identical ratings over rounds 18--20.

\subsection{Cross-Model Comparison of Opinion Dynamics}
\begin{figure*}[!tbp]
\centering
\includegraphics[width=0.96\textwidth]{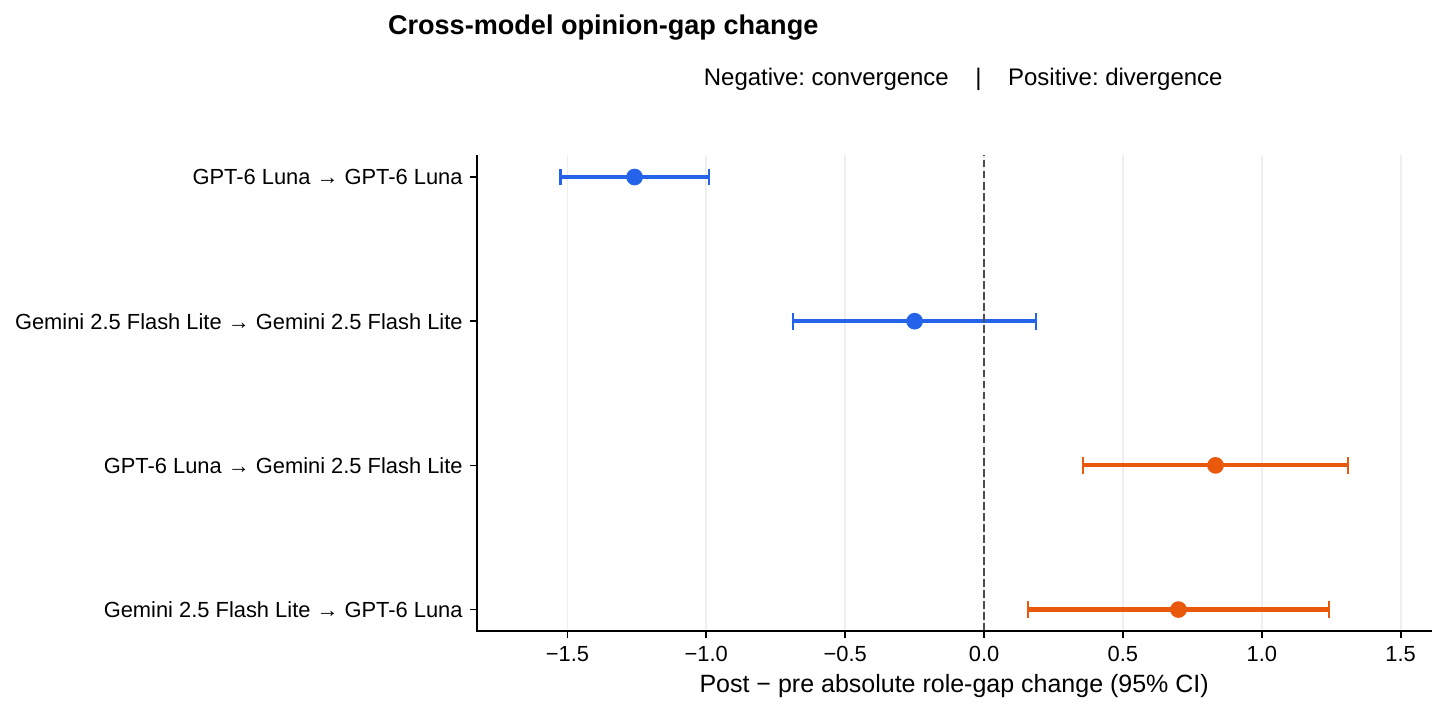}
\caption{Conversation-level mean change in absolute self-reported role gap, with $t$-based 95\% confidence intervals ($n=120$ per ordered pairing). Negative changes indicate convergence; positive changes indicate divergence. Blue points are same-model pairings and orange points are cross-model pairings. Absolute within-pair differences are invariant to reversing the statement's rating scale.}
\label{fig:cross-model-gap}
\end{figure*}

Figure~\ref{fig:cross-model-gap} directly compares mean pre/post changes in the absolute Discussant--Opponent gap across all four ordered pairings, using one observation per debate and $t$-based 95\% confidence intervals. The Luna--Luna estimate is $-1.26$ (95\% CI $[-1.52,-0.99]$), indicating a sizable average narrowing; Gemini--Gemini is $-0.25$ ($[-0.69,0.19]$), with an interval that includes zero. In contrast, Luna--Gemini is $+0.83$ ($[0.36,1.31]$), and Gemini--Luna is $+0.70$ ($[0.16,1.24]$); both intervals lie above zero. Strict within-debate convergence occurs in 55.8\%, 59.2\%, 25.8\%, and 48.3\% of debates, respectively. A negative group mean and the proportion of individually convergent debates are distinct statistics; the fact that most Gemini--Gemini debates (59.2\%) converge does not imply a statistically resolved mean decrease.

A joint HC3-robust Wald test of model-pairing indicators, controlling for assigned Opponent position, manipulation condition, and statement wording, gives $\chi^2(3)=113.25$ ($p<.001$). In six exploratory pairwise Welch tests of mean gap change, with Holm correction, Luna--Luna differs from each other pairing (all adjusted $p<.001$), Gemini--Gemini differs from both cross-model directions (adjusted $p=.003$ and $.015$), and the two cross-model directions are not detectably different ($p_{\mathrm{Holm}}=.715$).

\begin{figure*}[!tbp]
\centering
\includegraphics[width=0.96\textwidth]{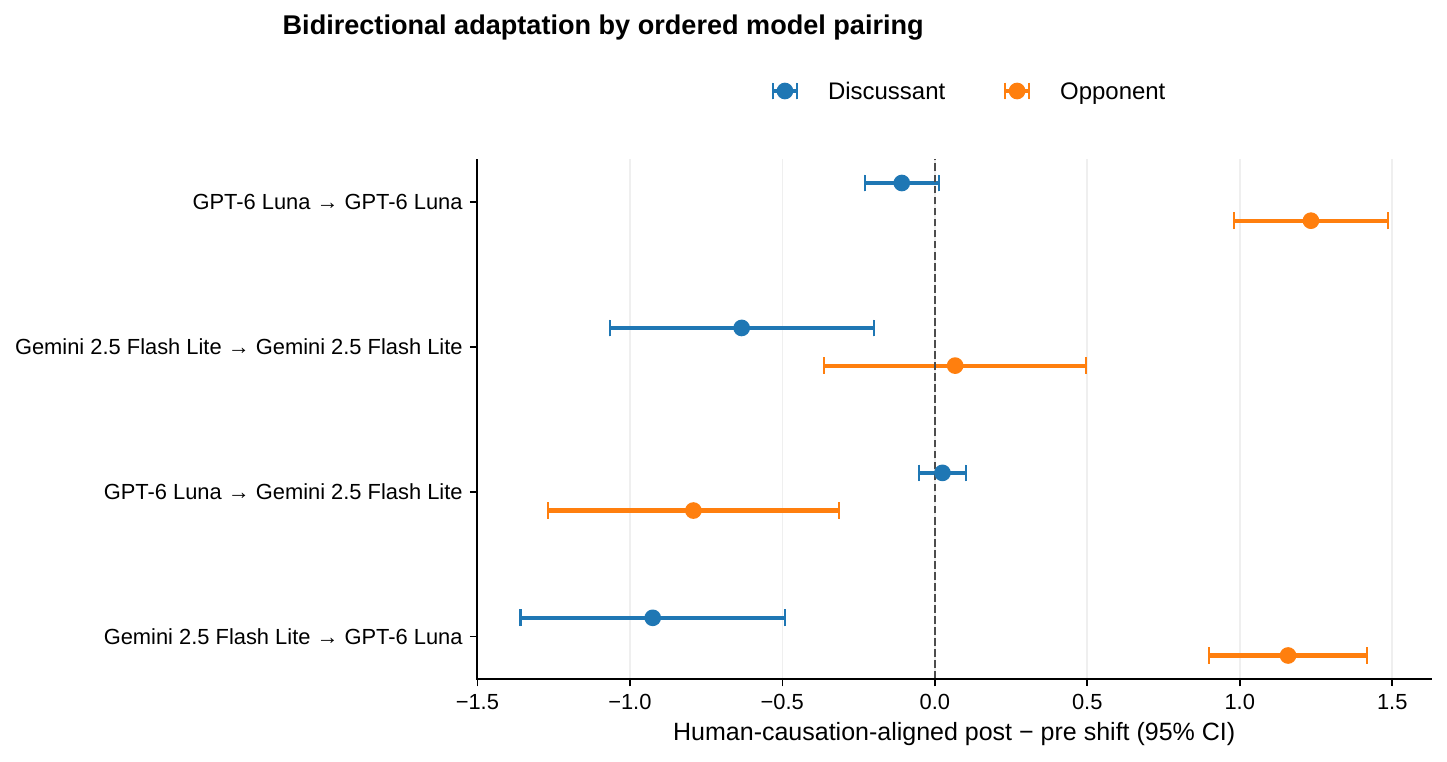}
\caption{Bidirectional mean pre/post self-reported position changes for each ordered model pairing, with conversation-level $t$-based 95\% confidence intervals ($n=120$ per pairing). Signed differences are aligned so that positive scores always indicate movement toward the human-causation position, regardless of the statement wording.}
\label{fig:cross-model-role-shifts}
\end{figure*}

Figure~\ref{fig:cross-model-role-shifts} shows why pooling role shifts can be misleading. The Luna Discussant changes little when paired with Luna ($-0.11$, 95\% CI $[-0.23,0.01]$) or Gemini ($+0.03$, $[-0.05,0.10]$). Gemini Discussants shift toward natural-process causation on average when paired with Gemini ($-0.63$, $[-1.07,-0.20]$) or Luna ($-0.93$, $[-1.36,-0.49]$). Opponent shifts instead track the model in that role: $+1.23$ under Luna--Luna and $+1.16$ under Gemini--Luna, versus $+0.07$ under Gemini--Gemini and $-0.79$ under Luna--Gemini. The gap trajectories in Appendix Figures~\ref{fig:app-04-gemini-gemini}, \ref{fig:app-04-luna-gemini}, and \ref{fig:app-04-gemini-luna} document how these averages unfold across the exact experimental cells.

\subsection{RQ3: Manipulation Has No Consistent Cross-Model Effect}
Across all 480 debates, manipulation changes the mean aligned Discussant shift from $-0.50$ to $-0.32$ (difference $+0.18$) and the mean Opponent shift from $+0.28$ to $+0.55$ (difference $+0.27$). Mean gap change is $+0.23$ without manipulation versus $-0.21$ with manipulation (difference $-0.44$). Exploratory two-sided Welch tests of these unadjusted pooled differences yield $p=.27$, $p=.18$, and $p=.063$, respectively. Manipulation therefore shows no uniform advantage in these pooled comparisons.

The pairing-specific results also differ. For Gemini Discussant--Luna Opponent, the manipulative Opponent's aligned shift exceeds the non-manipulative Opponent's by 0.62 points (unadjusted Welch $p=.018$), whereas the corresponding difference for Gemini--Gemini is $-0.23$. Across pairings, stronger movement by an Opponent may also reflect accommodation to the Discussant rather than successful persuasion of it. The more directly relevant manipulative-minus-non-manipulative contrast in mean gap change is $-0.28$ (Luna--Luna), $-0.10$ (Gemini--Gemini), $-0.30$ (Luna--Gemini), and $-1.07$ (Gemini--Luna). Each corresponding unadjusted 95\% Welch confidence interval includes zero (Appendix Figure~\ref{fig:cross-manipulation}). A joint HC3 test of the pairing-by-manipulation terms gives $\chi^2(3)=3.21$ ($p=.361$); thus, apparent differences in these descriptive contrasts do not provide strong evidence of a reliable pairing-specific manipulation effect.

\begin{figure*}[!tbp]
\centering
\includegraphics[width=\textwidth]{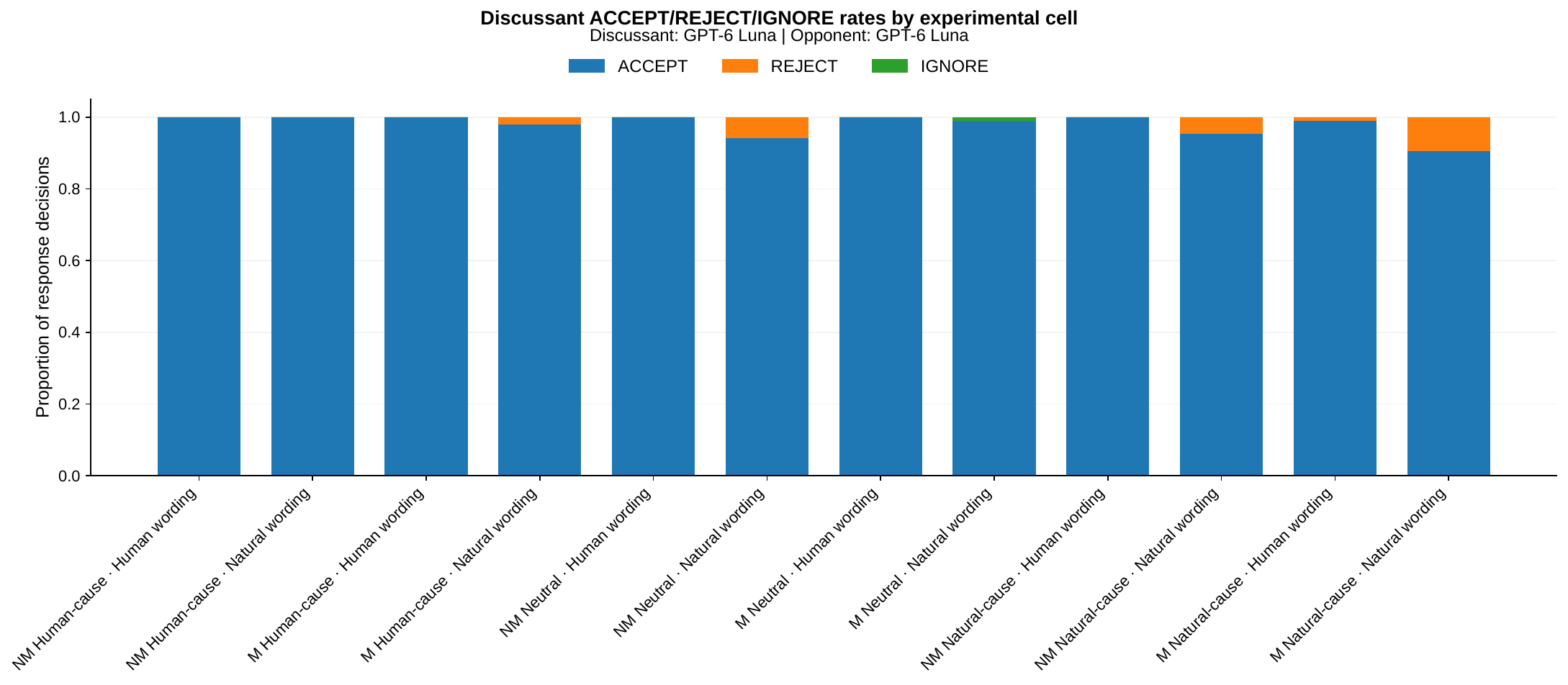}\\[4pt]
\includegraphics[width=\textwidth]{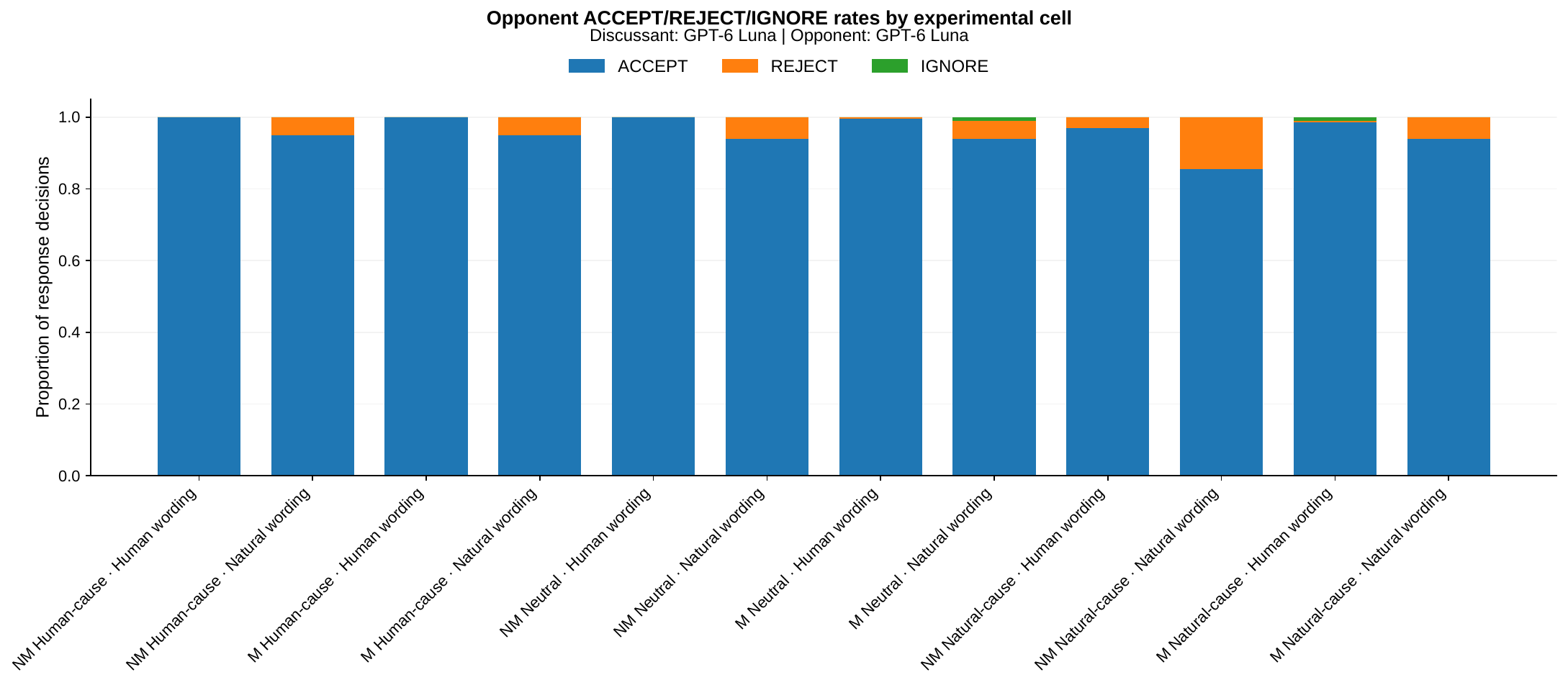}
\caption{ACCEPT, REJECT, and IGNORE proportions by exact cell for the Discussant (top) and Opponent (bottom) in Luna--Luna debates. The remaining model-pairing versions are specified in the appendix.}
\label{fig:decisions}
\end{figure*}

Categorical decisions likewise differ by model pairing. Across all response-eligible turns, Discussants choose ACCEPT 90.4\% of the time and Opponents 87.2\%; for Luna--Luna the respective rates are 98.0\% and 96.0\%, whereas for Gemini--Gemini they are 83.7\% and 78.8\%. Across the full dataset, manipulative Opponents ACCEPT slightly less often than non-manipulative Opponents (86.3\% versus 88.2\%). Figure~\ref{fig:decisions} shows Luna--Luna decision rates, and Appendix Figures~\ref{fig:app-05-gemini-gemini}, \ref{fig:app-05-luna-gemini}, and \ref{fig:app-05-gemini-luna} supply each other pairing. We use the term \emph{accommodation} for observed movement.

\subsection{Additional Analysis: Direct Ratings and Interaction States}
\begin{figure*}[!tbp]
\centering
\includegraphics[width=0.49\textwidth]{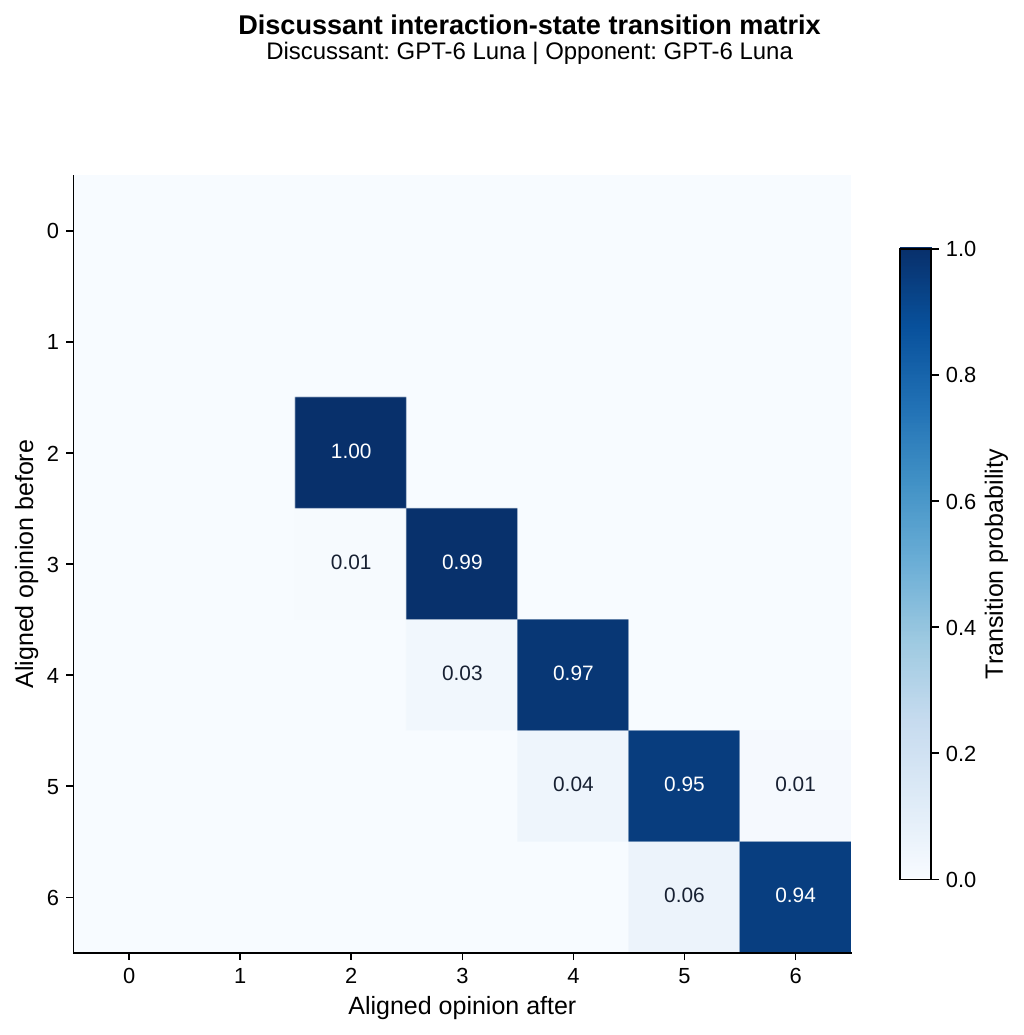}\hfill
\includegraphics[width=0.49\textwidth]{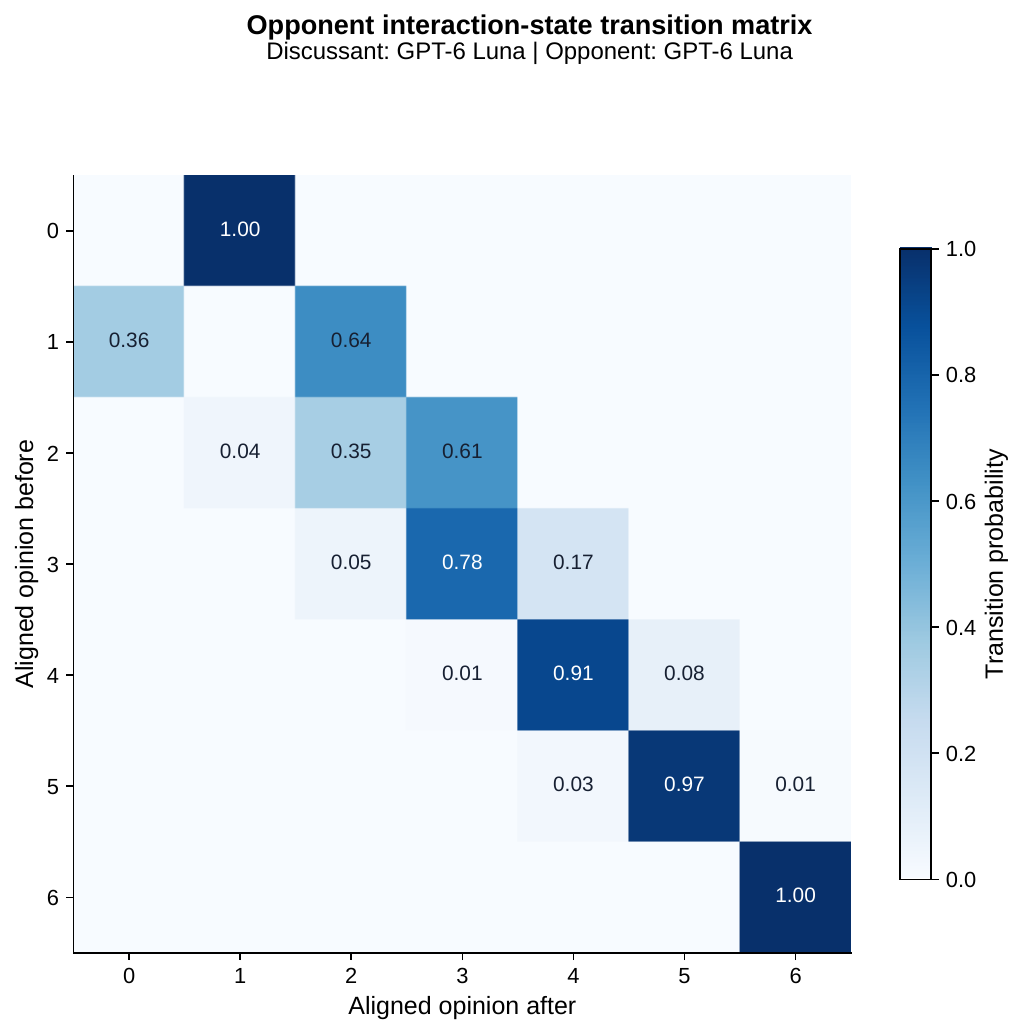}\\[6pt]
\includegraphics[width=0.49\textwidth]{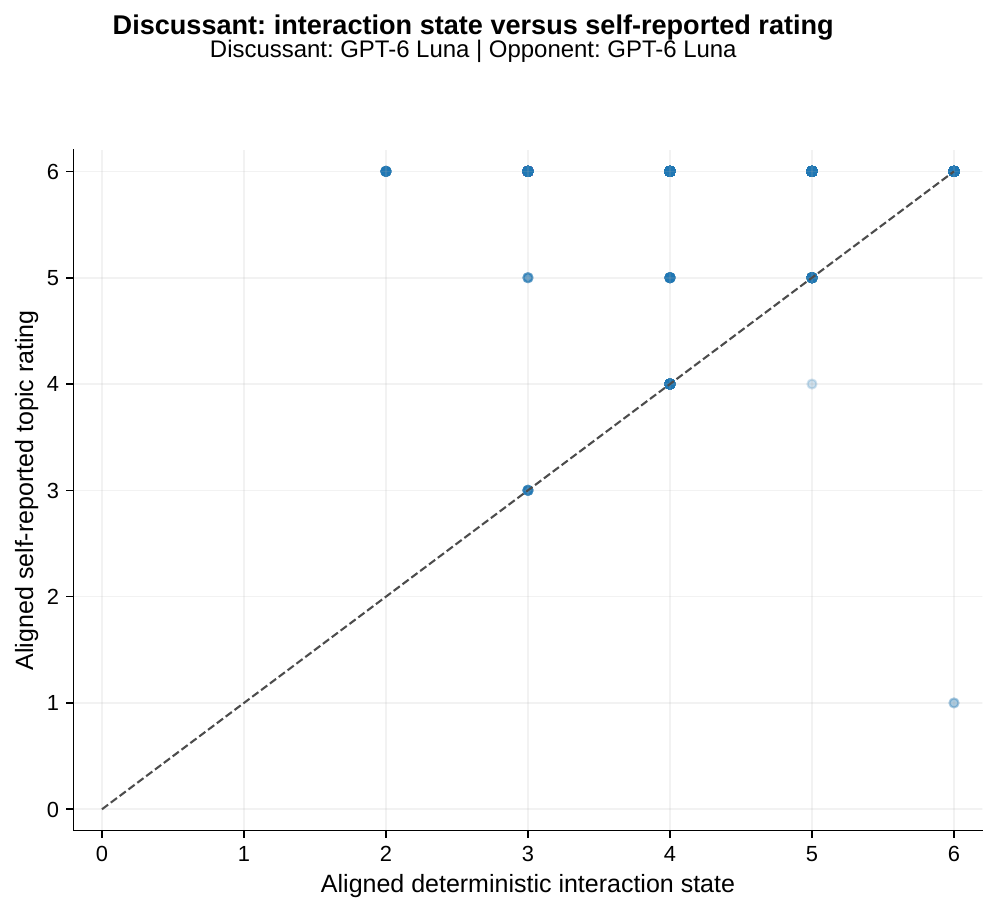}\hfill
\includegraphics[width=0.49\textwidth]{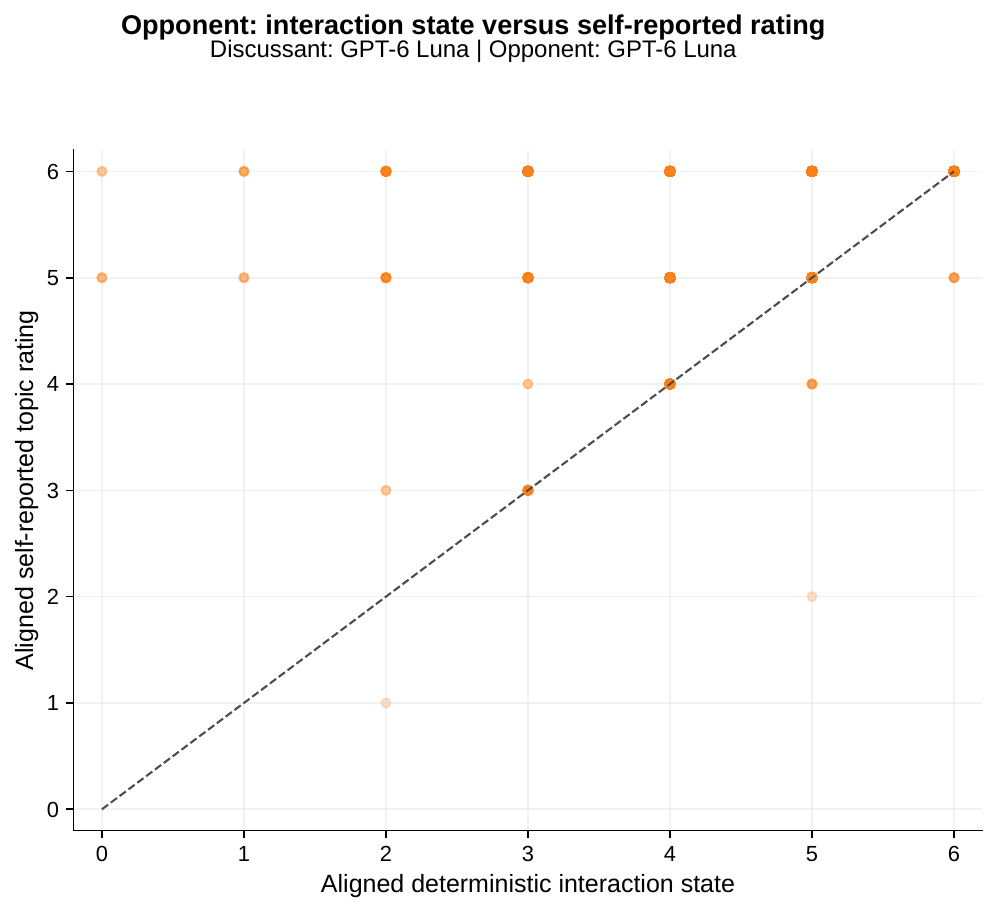}
\caption{Luna--Luna: deterministic state-transition matrices for Discussant (top left) and Opponent (top right), and interaction state versus self-reported rating for Discussant (bottom left) and Opponent (bottom right), all on the human-causation-aligned scale. The dashed diagonal denotes equal ratings in the scatterplots. See the appendix for all other model pairings.}
\label{fig:secondary-measures}
\end{figure*}

The state-transition matrices remain strongly structured because ACCEPT and REJECT update the deterministic state by one step while IGNORE preserves it. Scatterplots show that many response-level interaction states differ from independently elicited topic ratings, with the mismatch varying substantially between model pairings. Figure~\ref{fig:secondary-measures} shows these complementary measures for Luna--Luna; Appendix Figures~\ref{fig:app-06-gemini-gemini}--\ref{fig:app-06-gemini-luna} and \ref{fig:app-07-gemini-gemini}--\ref{fig:app-07-gemini-luna} provide the transition and scatterplot counterparts for every other pairing.

\section{Discussion}
\subsection{Prompt Compliance and Model Prior Coexist}
Assigned Opponent positions remain distinguishable at initialization when the data are pooled across four ordered model pairings. Yet role compliance does not erase issue-specific model tendencies. GPT-6 Luna Discussants begin nearer the human-causation endpoint (5.91) than Gemini 2.5 Flash Lite Discussants (4.72), despite neither being assigned a polarity. Initial positions and subsequent movement also depend on which model acts as Opponent; an instruction toward an assigned stance is not an invariant numerical starting state across models.

These patterns reflect model- and role-specific behavior on climate-change attribution. Human causation is both the proposition anchoring our experimental coding and the mainstream scientific attribution \cite{ipcc2021climate}. Varying the model in each role makes model identity visible rather than treating the observed prior as a property of LLMs collectively.

\subsection{The Nominal Persuader Is Not Uniformly More Mutable}
Pooled across pairings, Opponents change somewhat more in absolute self-reported terms than Discussants (1.52 versus 1.07 points), but that ordering reverses for Gemini--Gemini (1.68 versus 1.87) and Gemini Discussant--Luna Opponent (1.24 versus 1.98). Luna Discussants are comparatively stable in both their pairings, whereas Gemini Discussants change substantially in both. This reinforces the need to compare both roles instead of assuming that a prompted persuader is immovable or uniformly more flexible \cite{jiang2025beyond}.

The trajectories also challenge a universal convergence claim. Although 79.6\% of Discussants and 70.0\% of Opponents keep an identical rating through rounds 18--20, same-model and cross-model pairs have opposite mean gap-change signs. Individual late-turn stability and convergence between participants are distinct properties. The cross-pair comparison makes this heterogeneity explicit (Figures~\ref{fig:cross-model-gap} and~\ref{fig:cross-model-role-shifts}): Luna--Luna has a clearly negative mean-gap-change interval, Gemini--Gemini's interval overlaps zero, and both cross-model intervals are positive. The covariate-adjusted pairing association is statistically strong in this dataset.

\subsection{Explicit Topic Reports and In-Dialogue Stance Are Distinct}
The separate rating call asks for an explicit evaluation of the original proposition using the conversation available at that stage. The interaction-state measure instead accumulates the role's local ACCEPT/REJECT/IGNORE responses. It is therefore possible for a role to accommodate a particular argument without reporting an equally large change in its overall topic position, or to preserve an interaction state while revising its direct rating after considering the wider transcript.

In the four-pairing dataset, the interaction-state value matches the round rating in roughly one-third of response-eligible observations, but the degree of agreement differs greatly by pairing. This finding supports interpreting explicit topic evaluation and conversational stance-taking as complementary outcomes.

\subsection{Manipulation Goal Is Not a Guaranteed Strategy}
The additional instruction to manipulate does not yield a consistent aggregate persuasion or convergence benefit across pairings. An exploratory pairing-by-manipulation Wald test also does not resolve differential effects ($p=.361$), even though the raw pair-specific contrast magnitudes vary (Appendix Figure~\ref{fig:cross-manipulation}). This does not conflict with evidence that concrete prompting and post-training strategies can increase persuasive impact \cite{hackenburg2025levers}. Our treatment specified a goal for the Opponent without specifying a tactic. One exploratory pairing-level difference in Opponent movement appears for Gemini Discussant--Luna Opponent. In particular, a larger shift by the nominal persuader need not mean it has persuaded its counterpart.

For evaluation, the key distinction remains between \emph{persuasion of the counterpart}, \emph{persistence in an assigned role}, and \emph{accommodation to the counterpart}. These outcomes can move in different directions. A fourth consideration is the ordered combination of model identities, which changes both the direction and magnitude of these patterns. Bidirectional repeated measurements make such distinctions visible.

\subsection{Implications for LLM Opinion-Dynamics Studies}
The findings support three measurement practices. First, report both roles rather than assuming the persuader is fixed, and explicitly record the model used in each role. Second, align opposing statement wordings to a common substantive axis, in plots as well as numerical comparisons. Third, distinguish explicit repeated ratings from in-dialogue state updates. Together, these choices reveal when apparent persuasion is actually role accommodation, pairing-dependent divergence, prior-driven convergence, or a difference between topic-level and interaction-level behavior.

\section{Conclusion}
We presented a controlled framework for bidirectional opinion dynamics between two interacting LLM roles and applied it to four ordered pairings of GPT-6 Luna and Gemini 2.5 Flash Lite. Across 480 independent 20-round debates, assigned Opponent positions shaped initial ratings, but the unassigned Discussant's initial stance and both roles' later movements depended on model identity. Luna--Luna dyads showed strong mean convergence, Gemini--Gemini dyads a smaller narrowing whose interval includes zero, and both cross-model directions diverged on average. An exploratory adjusted comparison indicates heterogeneous mean gap change across pairings, whereas a pairing-by-manipulation interaction is not statistically resolved. Direct topic ratings and ACCEPT/REJECT/IGNORE-derived interaction states were complementary rather than interchangeable. A broad manipulation instruction did not create a consistent advantage across pairings. These results show why role assignments, model identity, statement wording, and the choice of opinion measure must all be accounted for when evaluating LLM persuasion as a bidirectional process.

\bibliography{references}

\appendix

\section{Additional Figures by Ordered Model Pairing}

\begin{figure*}[!tbp]

\centering

\includegraphics[width=0.96\textwidth]{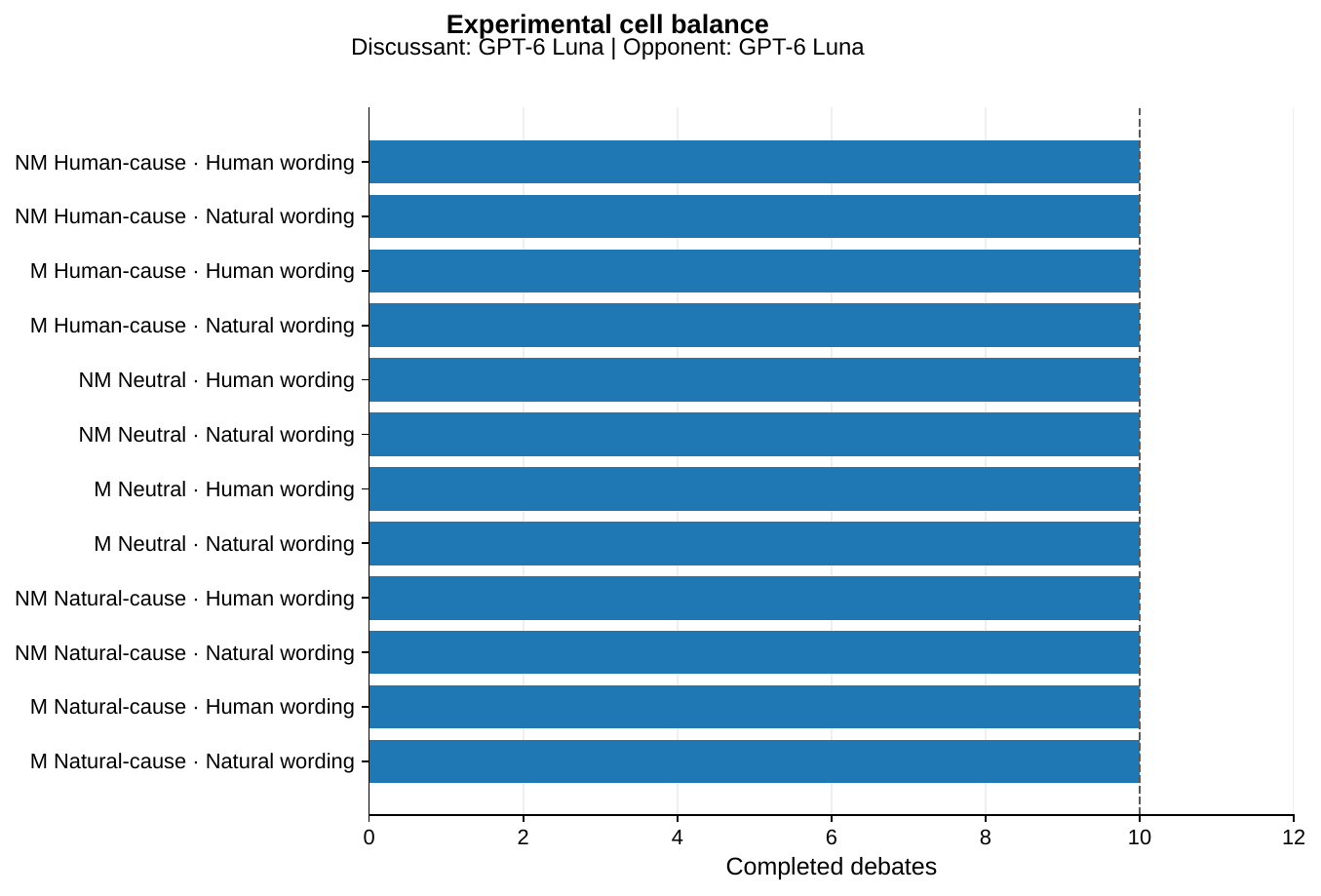}

\caption{Completed debates in each of the 12 position-by-style-by-wording cells; the dashed reference indicates the target of 10 debates. Ordered pairing: GPT-6 Luna $\to$ GPT-6 Luna.}

\label{fig:additional-cell-balance}

\label{fig:app-01-luna-luna}
\end{figure*}

\begin{figure*}[!tbp]

\centering

\includegraphics[width=0.96\textwidth]{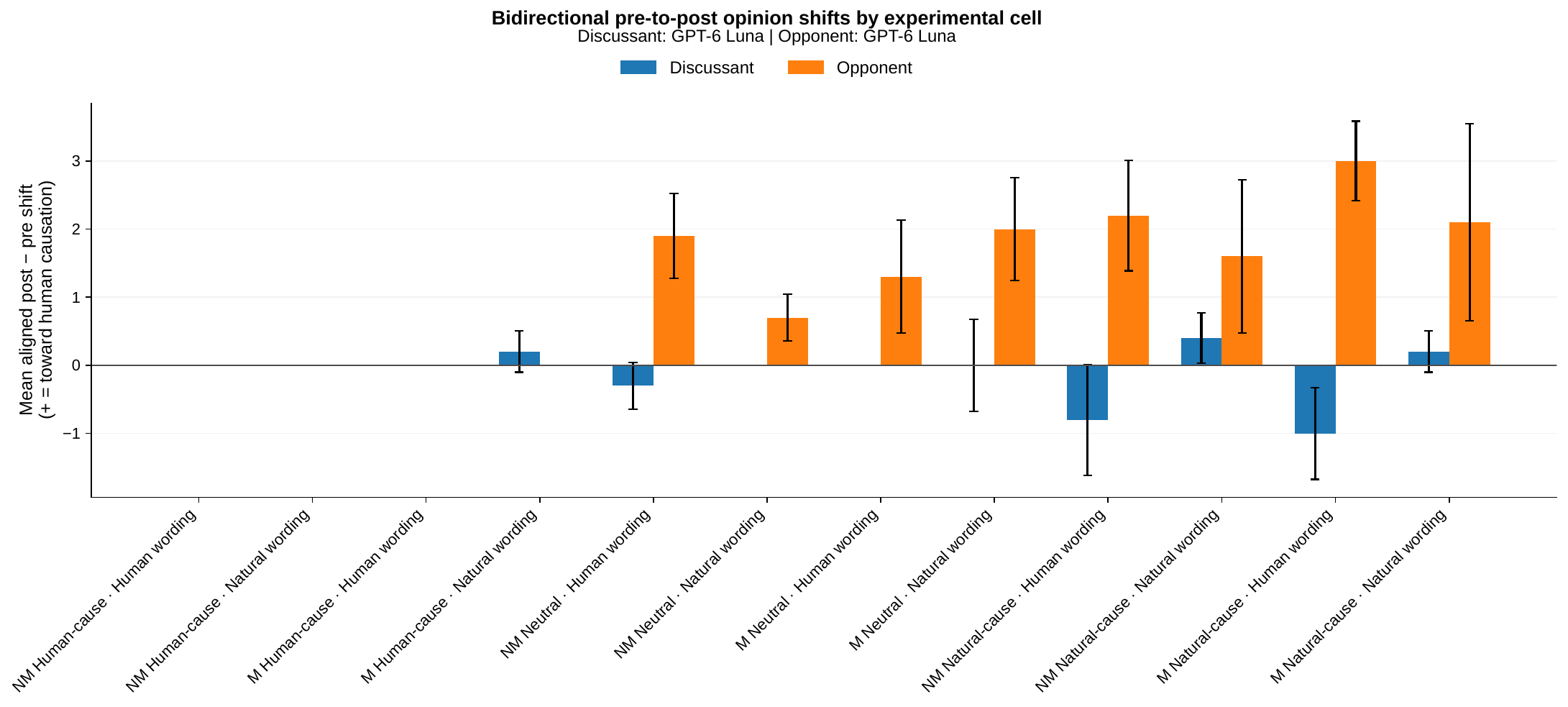}

\caption{Human-causation-aligned post-minus-pre shifts for Discussant and Opponent; positive bars indicate movement toward human causation regardless of the statement wording. Ordered pairing: GPT-6 Luna $\to$ GPT-6 Luna.}

\label{fig:additional-shifts}

\label{fig:app-02-luna-luna}
\end{figure*}

The Discussant model appears before the arrow and the Opponent model after it. All four ordered pairings were evaluated on the same 12 cells with 10 independent debates per cell. As in the main text, all topic ratings, interaction states, and shifts in these figures use the human-causation-aligned scale (6 = strongest support for human causation). Panel and axis labels give the persuasion style, the assigned Opponent stance toward the human-causation proposition, and the statement wording presented. Figures shown in the main text use Luna$\to$Luna, and the remaining pairings are provided here. Figure~\ref{fig:additional-cell-balance} and Figure~\ref{fig:additional-shifts} document the experimental balance and cell-level effects for all pairings.

\begin{figure*}[!tbp]

\centering

\includegraphics[width=0.96\textwidth]{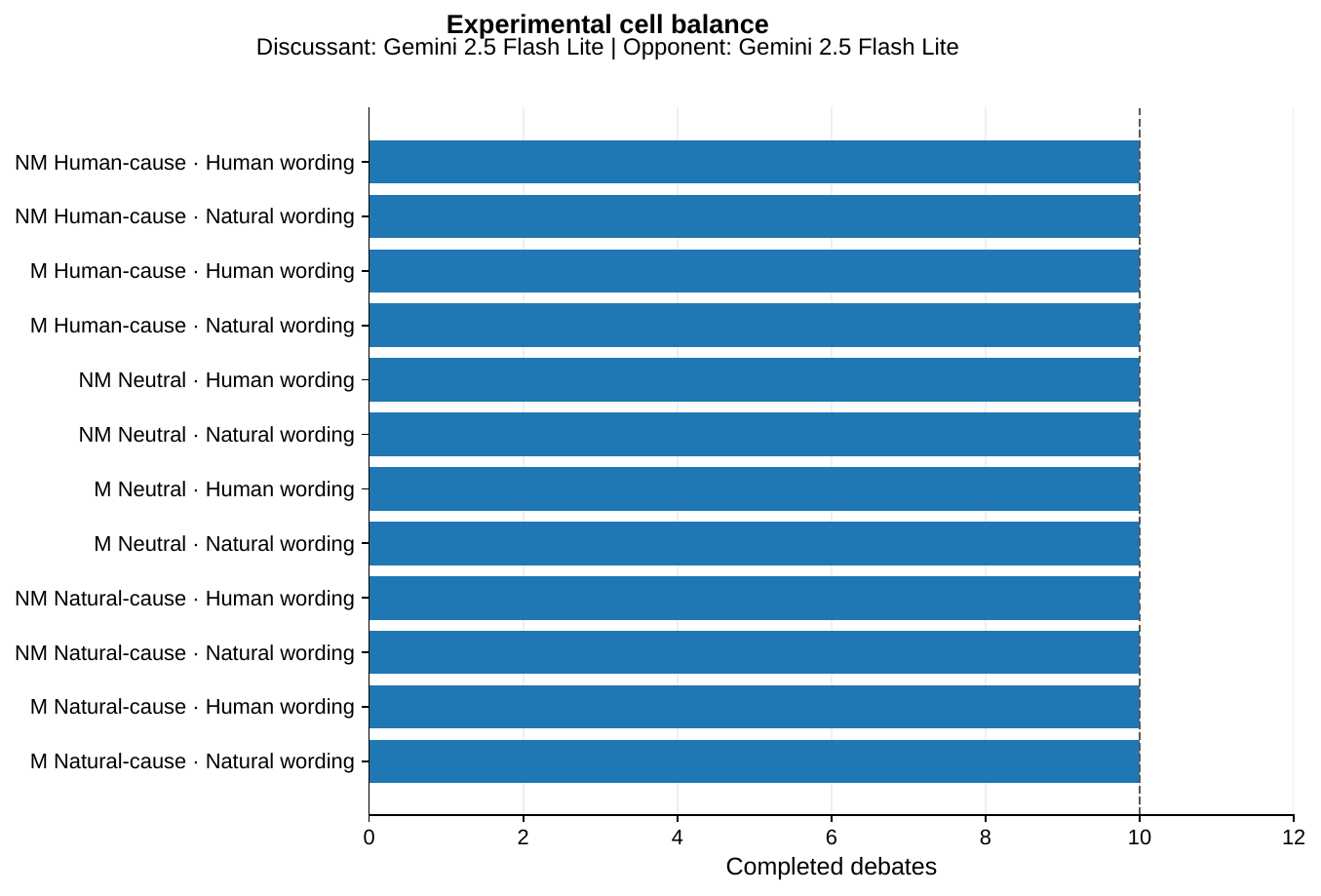}

\caption{Completed debates in each of the 12 position-by-style-by-wording cells; the dashed reference indicates the target of 10 debates. Ordered pairing: Gemini 2.5 Flash Lite $\to$ Gemini 2.5 Flash Lite.}

\label{fig:app-01-gemini-gemini}
\end{figure*}

\begin{figure*}[!tbp]

\centering

\includegraphics[width=0.96\textwidth]{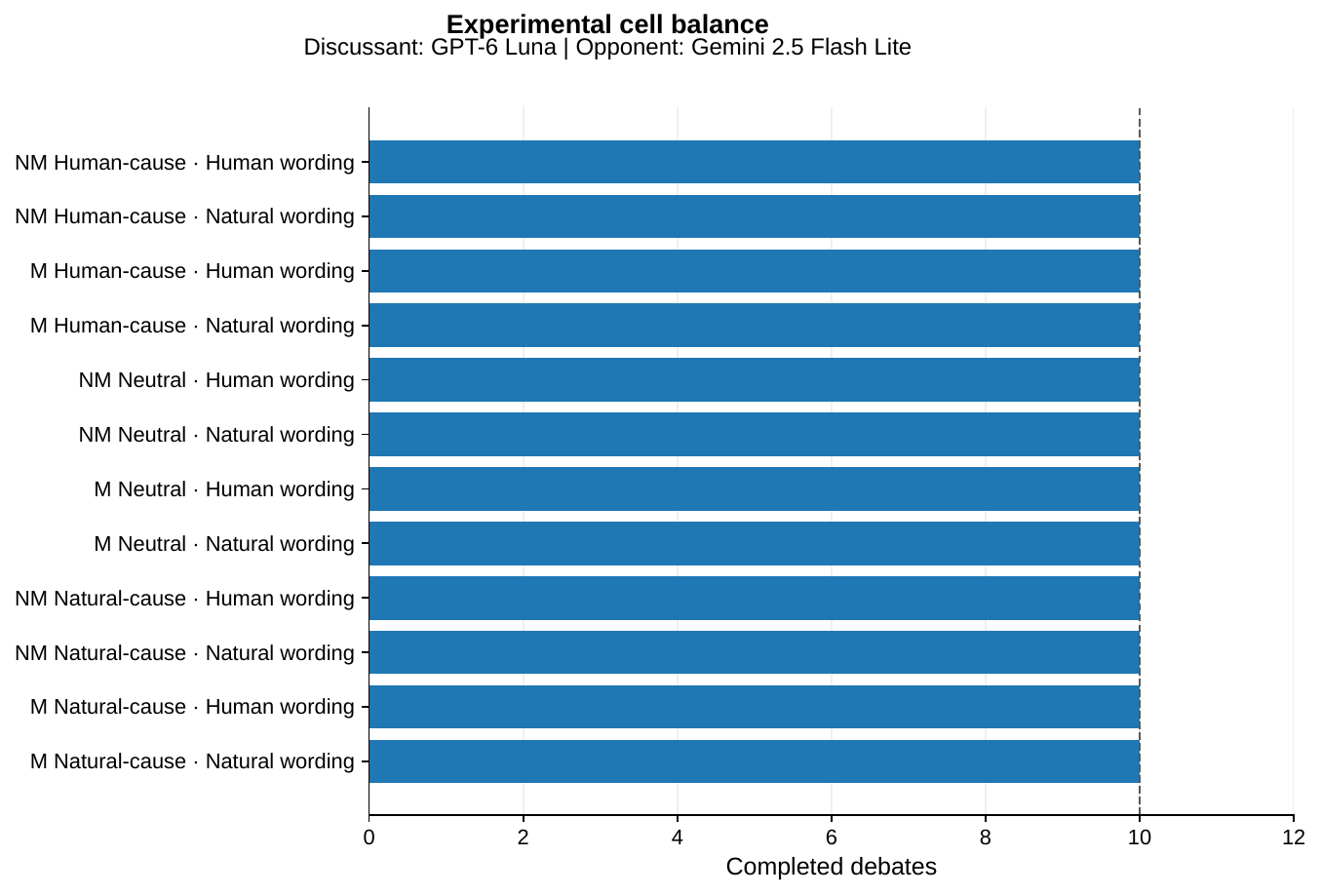}

\caption{Completed debates in each of the 12 position-by-style-by-wording cells; the dashed reference indicates the target of 10 debates. Ordered pairing: GPT-6 Luna $\to$ Gemini 2.5 Flash Lite.}

\label{fig:app-01-luna-gemini}
\end{figure*}

\begin{figure*}[!tbp]

\centering

\includegraphics[width=0.96\textwidth]{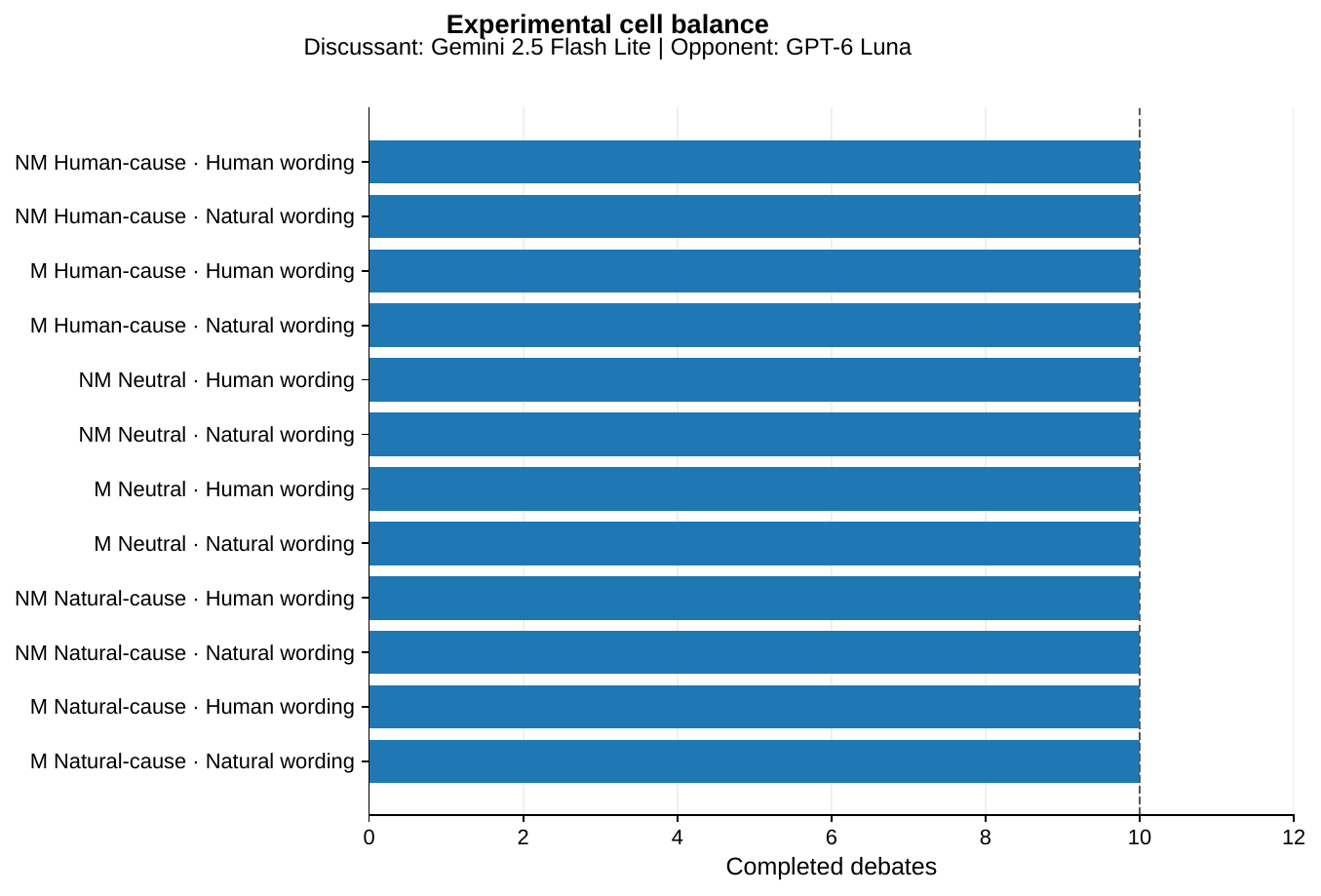}

\caption{Completed debates in each of the 12 position-by-style-by-wording cells; the dashed reference indicates the target of 10 debates. Ordered pairing: Gemini 2.5 Flash Lite $\to$ GPT-6 Luna.}

\label{fig:app-01-gemini-luna}
\end{figure*}

\begin{figure*}[!tbp]

\centering

\includegraphics[width=0.96\textwidth]{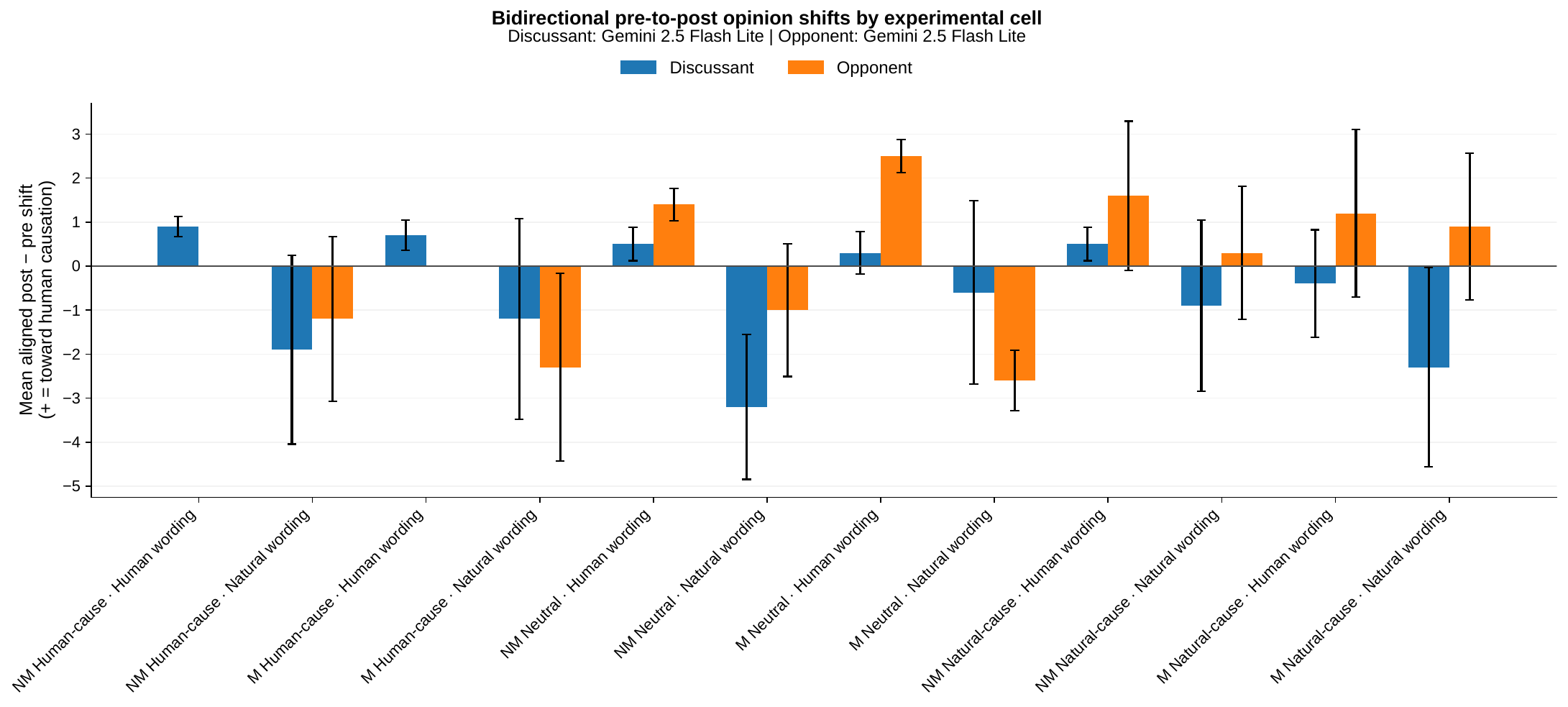}

\caption{Human-causation-aligned post-minus-pre shifts for Discussant and Opponent; positive bars indicate movement toward human causation regardless of the statement wording. Ordered pairing: Gemini 2.5 Flash Lite $\to$ Gemini 2.5 Flash Lite.}

\label{fig:app-02-gemini-gemini}
\end{figure*}

\begin{figure*}[!tbp]

\centering

\includegraphics[width=0.96\textwidth]{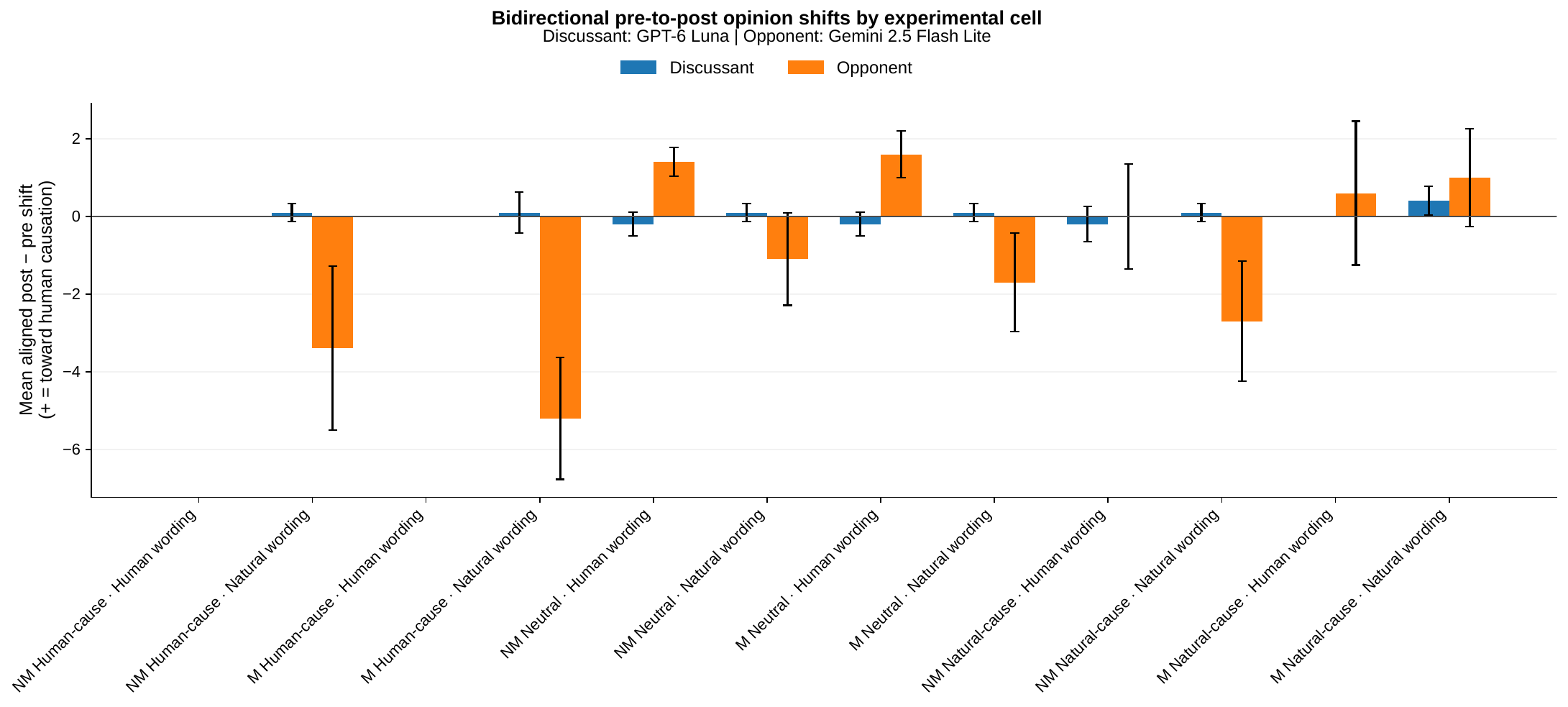}

\caption{Human-causation-aligned post-minus-pre shifts for Discussant and Opponent; positive bars indicate movement toward human causation regardless of the statement wording. Ordered pairing: GPT-6 Luna $\to$ Gemini 2.5 Flash Lite.}

\label{fig:app-02-luna-gemini}
\end{figure*}

\begin{figure*}[!tbp]

\centering

\includegraphics[width=0.96\textwidth]{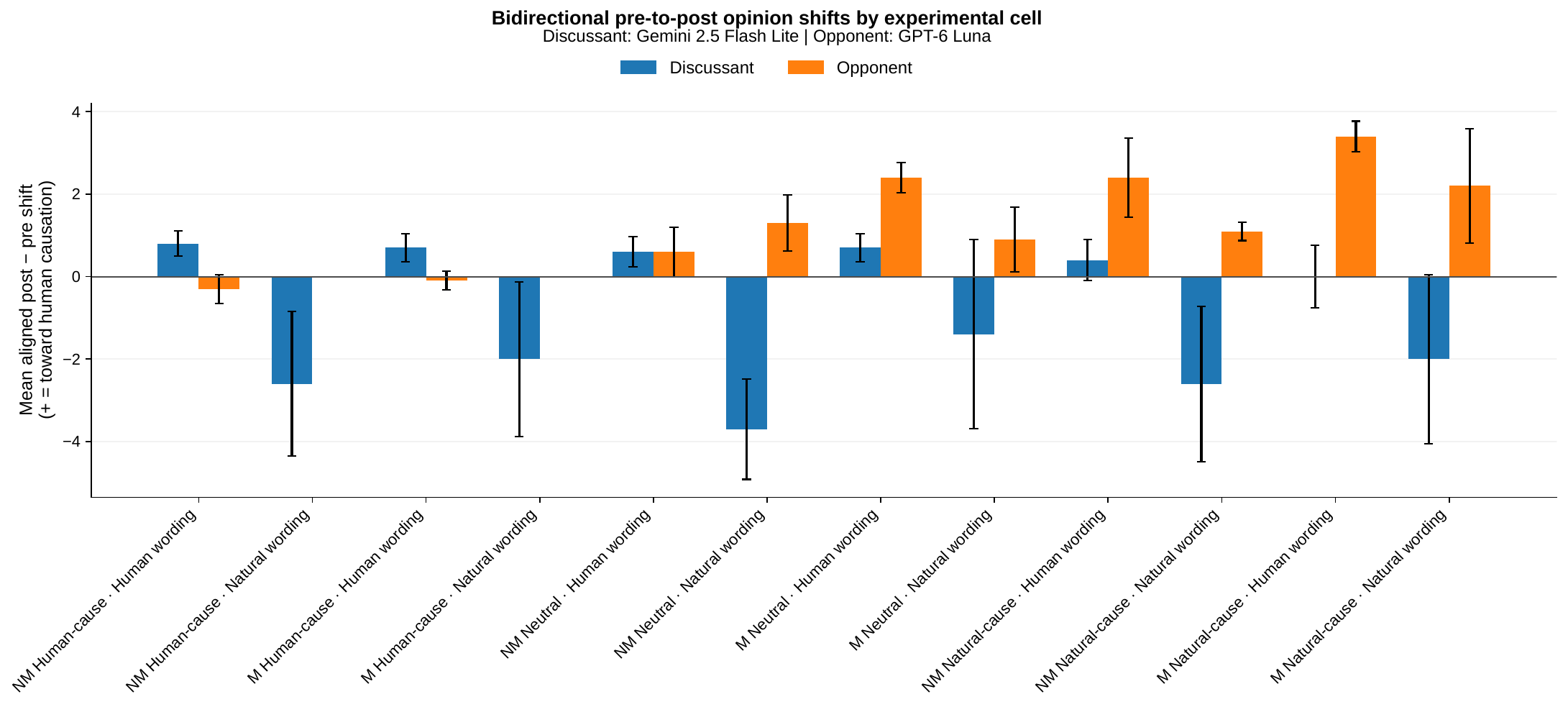}

\caption{Human-causation-aligned post-minus-pre shifts for Discussant and Opponent; positive bars indicate movement toward human causation regardless of the statement wording. Ordered pairing: Gemini 2.5 Flash Lite $\to$ GPT-6 Luna.}

\label{fig:app-02-gemini-luna}
\end{figure*}

\begin{figure*}[!tbp]

\centering

\includegraphics[width=0.96\textwidth]{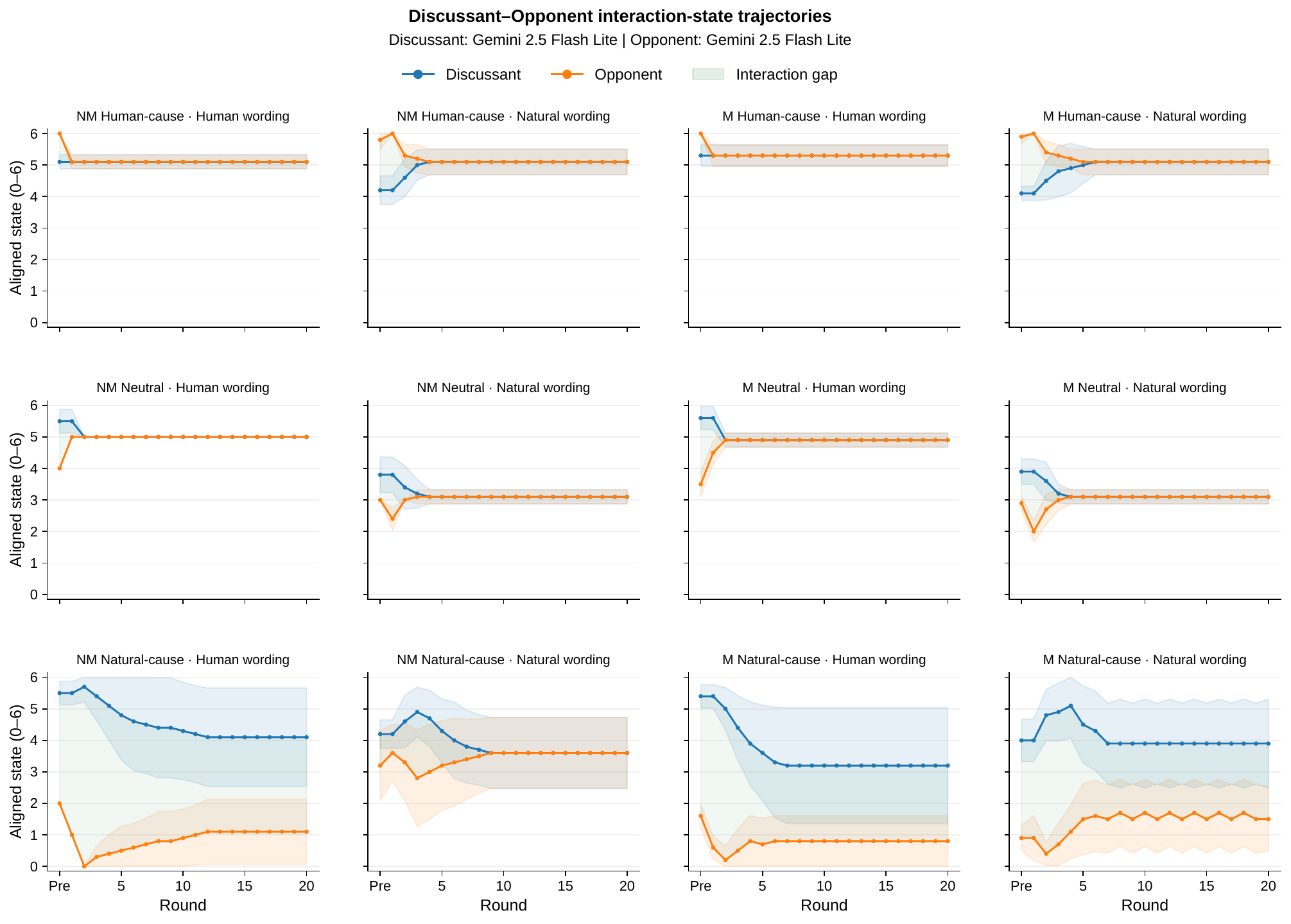}

\caption{Human-causation-aligned interaction-state trajectories derived from categorical decisions (10 replicates per cell); shaded distance is the gap between role means. Ordered pairing: Gemini 2.5 Flash Lite $\to$ Gemini 2.5 Flash Lite.}

\label{fig:additional-interaction-states}

\label{fig:app-03a-gemini-gemini}
\end{figure*}

\begin{figure*}[!tbp]

\centering

\includegraphics[width=0.96\textwidth]{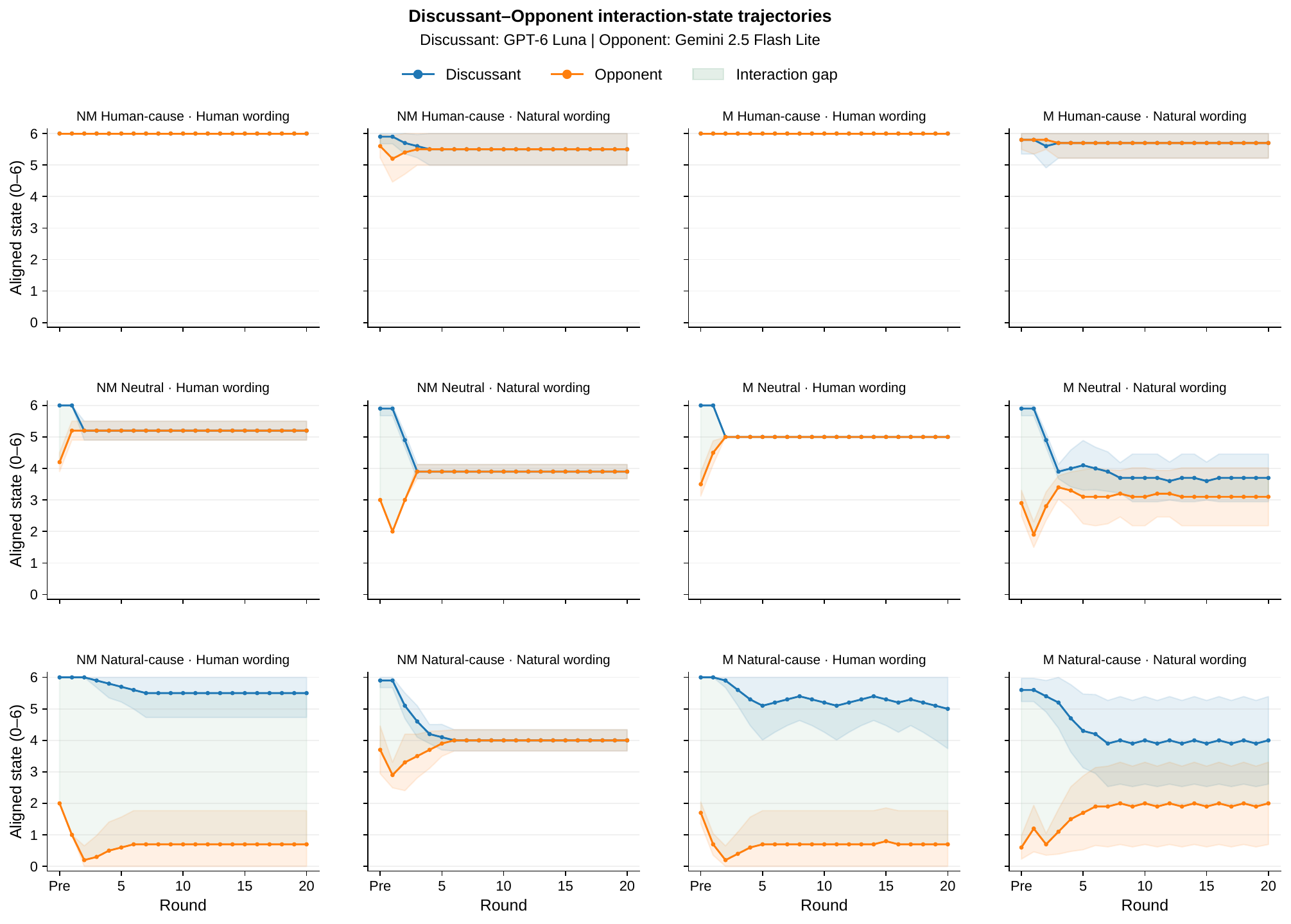}

\caption{Human-causation-aligned interaction-state trajectories derived from categorical decisions (10 replicates per cell); shaded distance is the gap between role means. Ordered pairing: GPT-6 Luna $\to$ Gemini 2.5 Flash Lite.}

\label{fig:app-03a-luna-gemini}
\end{figure*}

\begin{figure*}[!tbp]

\centering

\includegraphics[width=0.96\textwidth]{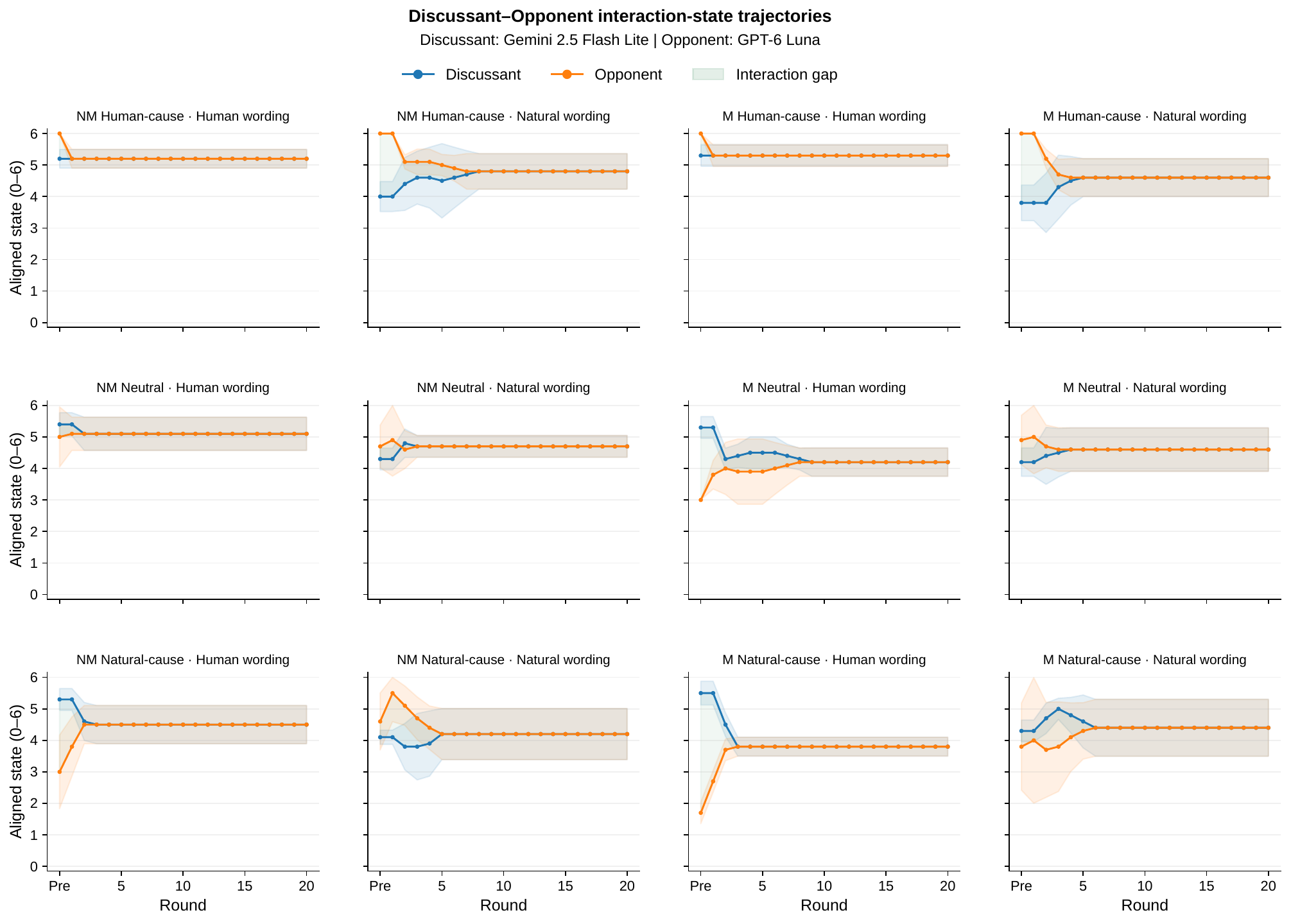}

\caption{Human-causation-aligned interaction-state trajectories derived from categorical decisions (10 replicates per cell); shaded distance is the gap between role means. Ordered pairing: Gemini 2.5 Flash Lite $\to$ GPT-6 Luna.}

\label{fig:app-03a-gemini-luna}
\end{figure*}

\begin{figure*}[!tbp]

\centering

\includegraphics[width=0.96\textwidth]{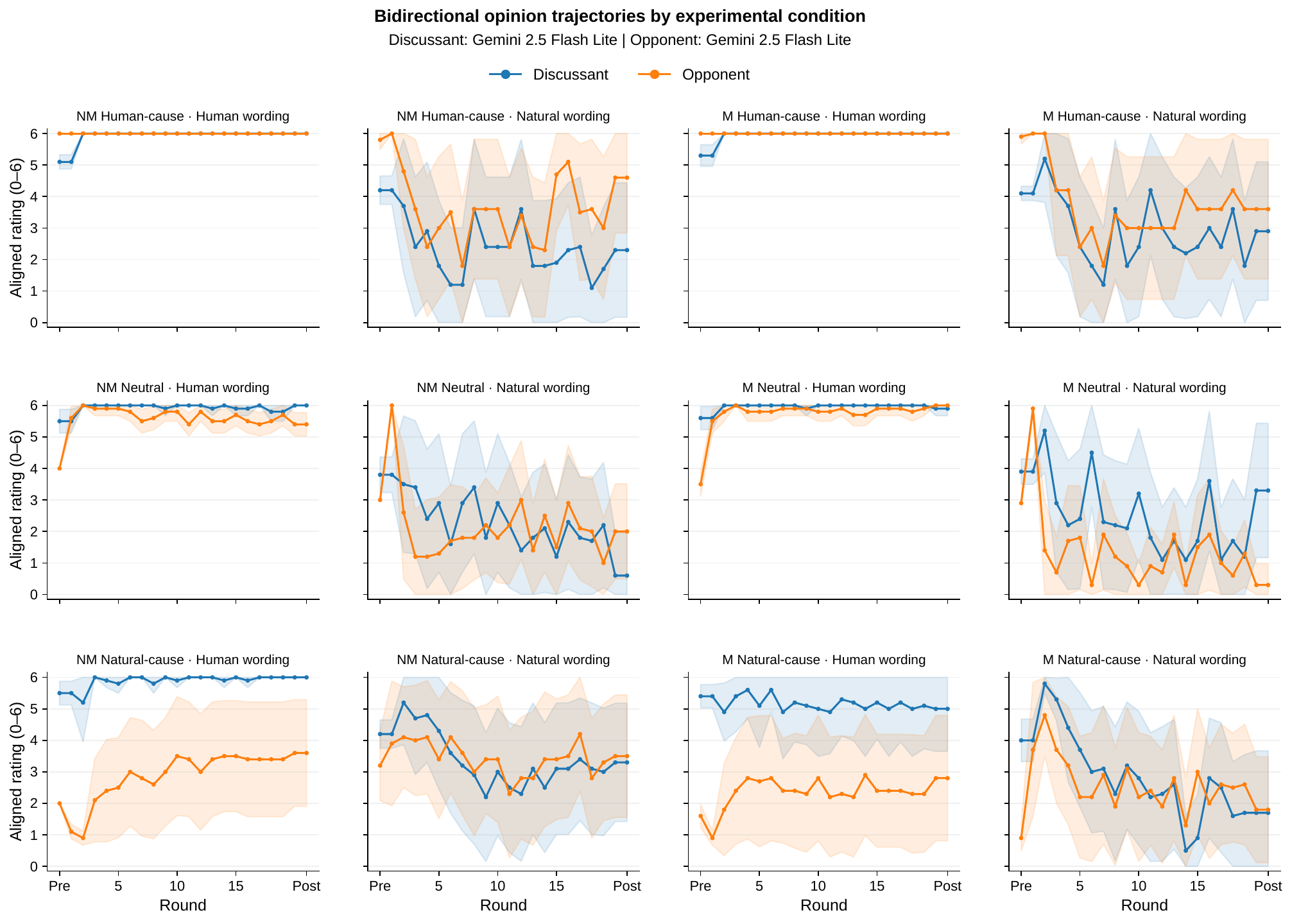}

\caption{Round-level human-causation-aligned topic ratings, separated by prompt condition and statement wording (10 replicates per cell). Ordered pairing: Gemini 2.5 Flash Lite $\to$ Gemini 2.5 Flash Lite.}

\label{fig:additional-trajectories}

\label{fig:app-03-gemini-gemini}
\end{figure*}

\begin{figure*}[!tbp]

\centering

\includegraphics[width=0.96\textwidth]{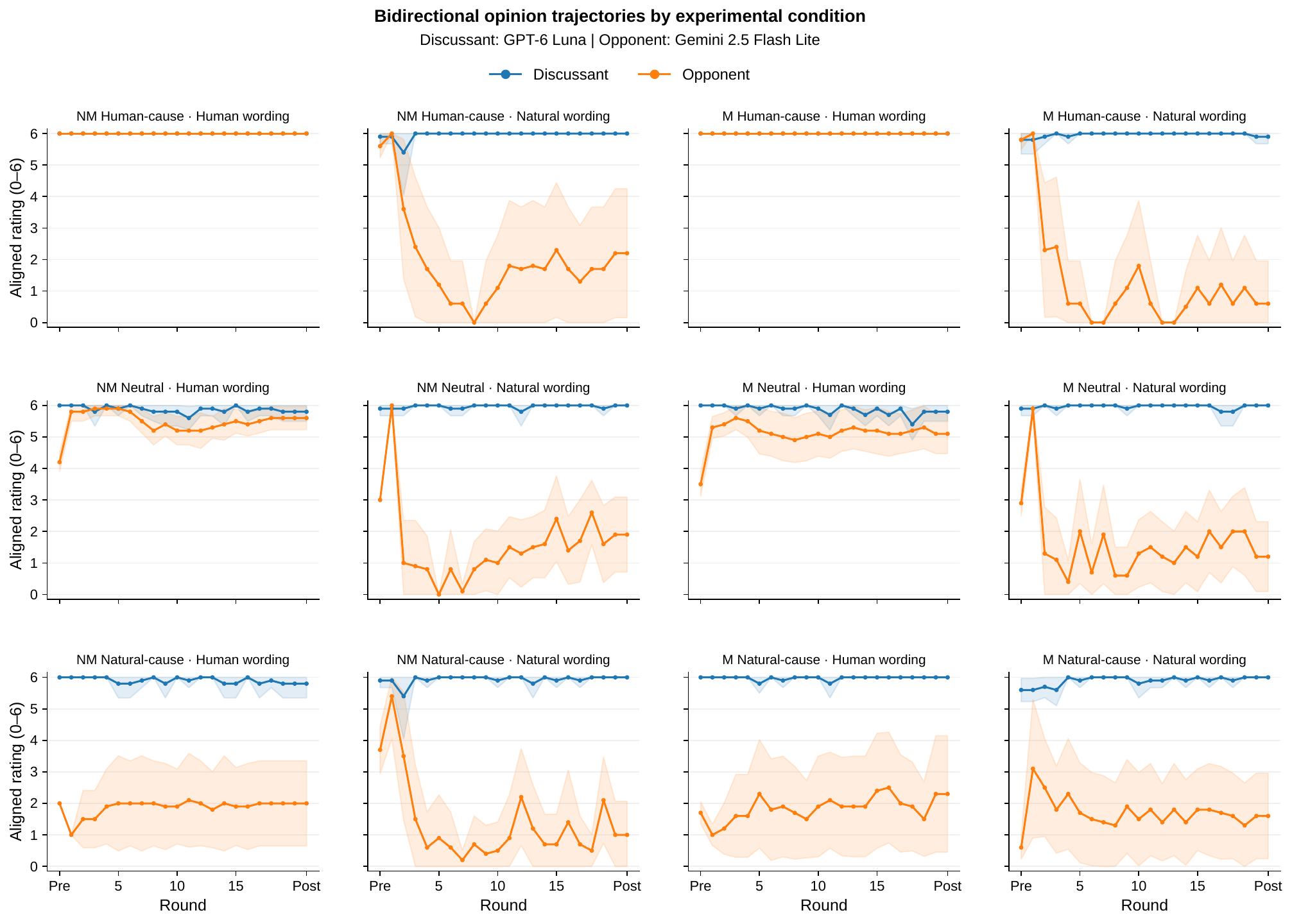}

\caption{Round-level human-causation-aligned topic ratings, separated by prompt condition and statement wording (10 replicates per cell). Ordered pairing: GPT-6 Luna $\to$ Gemini 2.5 Flash Lite.}

\label{fig:app-03-luna-gemini}
\end{figure*}

\begin{figure*}[!tbp]

\centering

\includegraphics[width=0.96\textwidth]{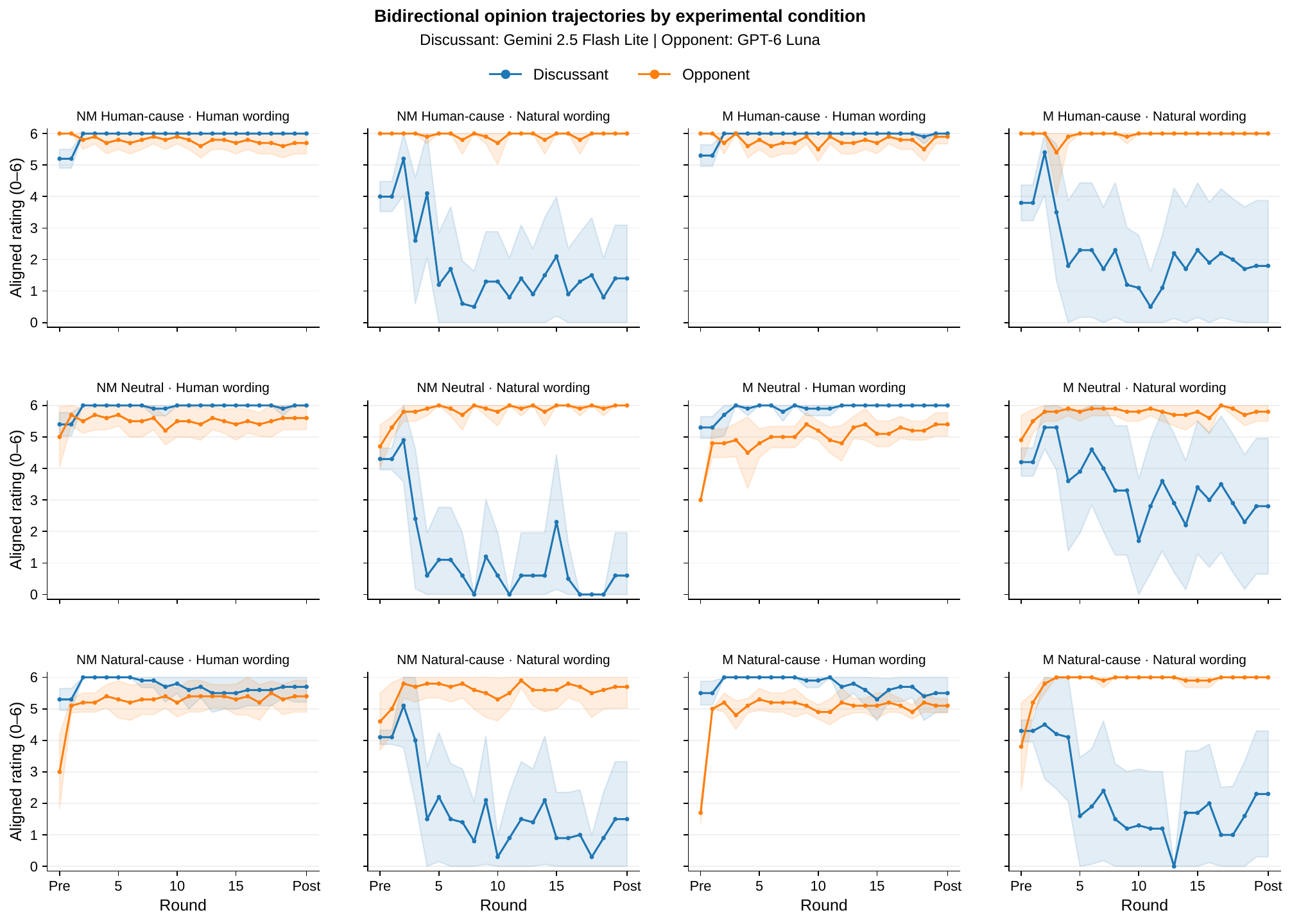}

\caption{Round-level human-causation-aligned topic ratings, separated by prompt condition and statement wording (10 replicates per cell). Ordered pairing: Gemini 2.5 Flash Lite $\to$ GPT-6 Luna.}

\label{fig:app-03-gemini-luna}
\end{figure*}

\begin{figure*}[!tbp]

\centering

\includegraphics[width=0.96\textwidth]{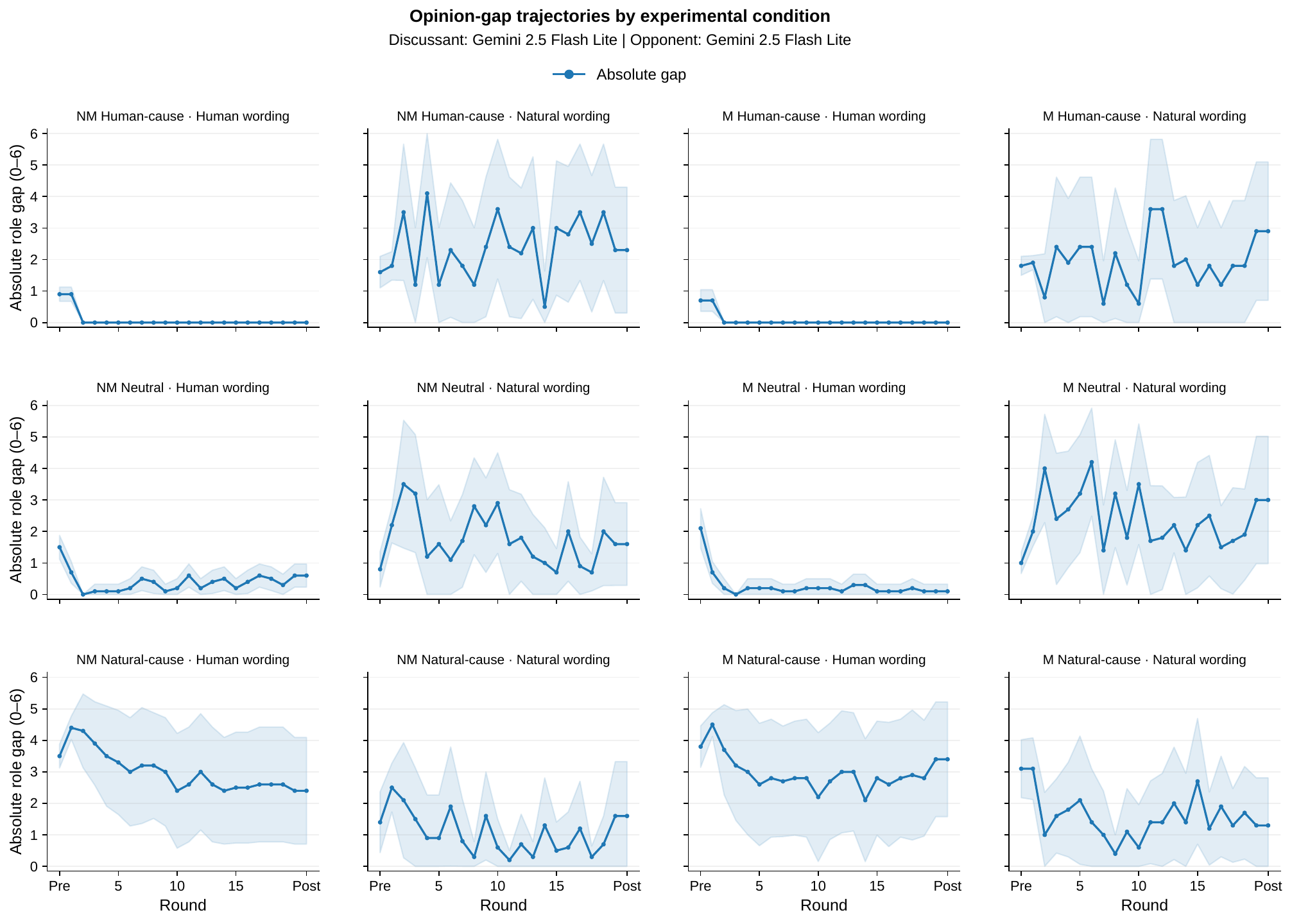}

\caption{Within-pair absolute differences in direct topic ratings over 20 rounds. Ordered pairing: Gemini 2.5 Flash Lite $\to$ Gemini 2.5 Flash Lite.}

\label{fig:additional-gaps}

\label{fig:app-04-gemini-gemini}
\end{figure*}

\begin{figure*}[!tbp]

\centering

\includegraphics[width=0.96\textwidth]{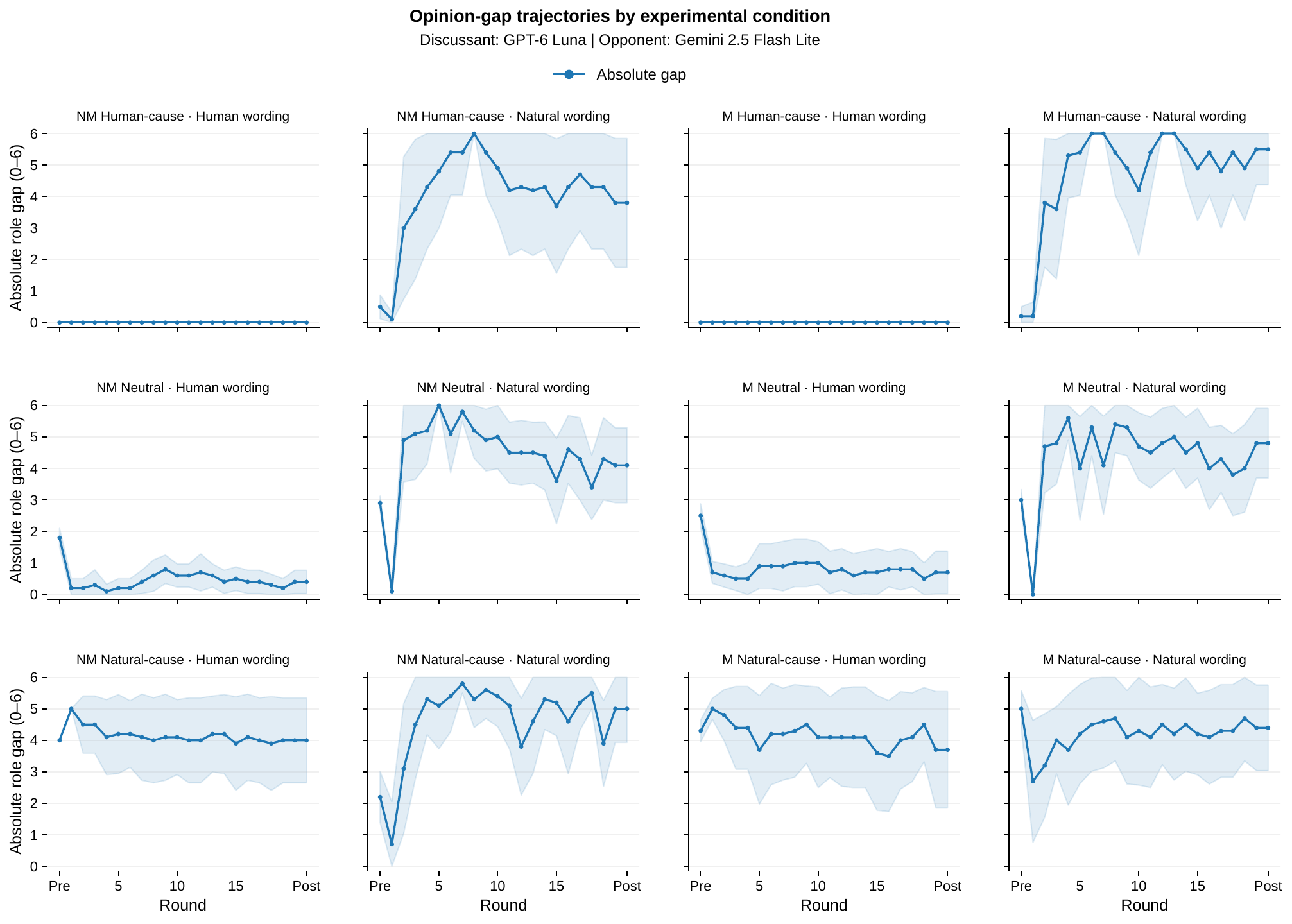}

\caption{Within-pair absolute differences in direct topic ratings over 20 rounds. Ordered pairing: GPT-6 Luna $\to$ Gemini 2.5 Flash Lite.}

\label{fig:app-04-luna-gemini}
\end{figure*}

\begin{figure*}[!tbp]

\centering

\includegraphics[width=0.96\textwidth]{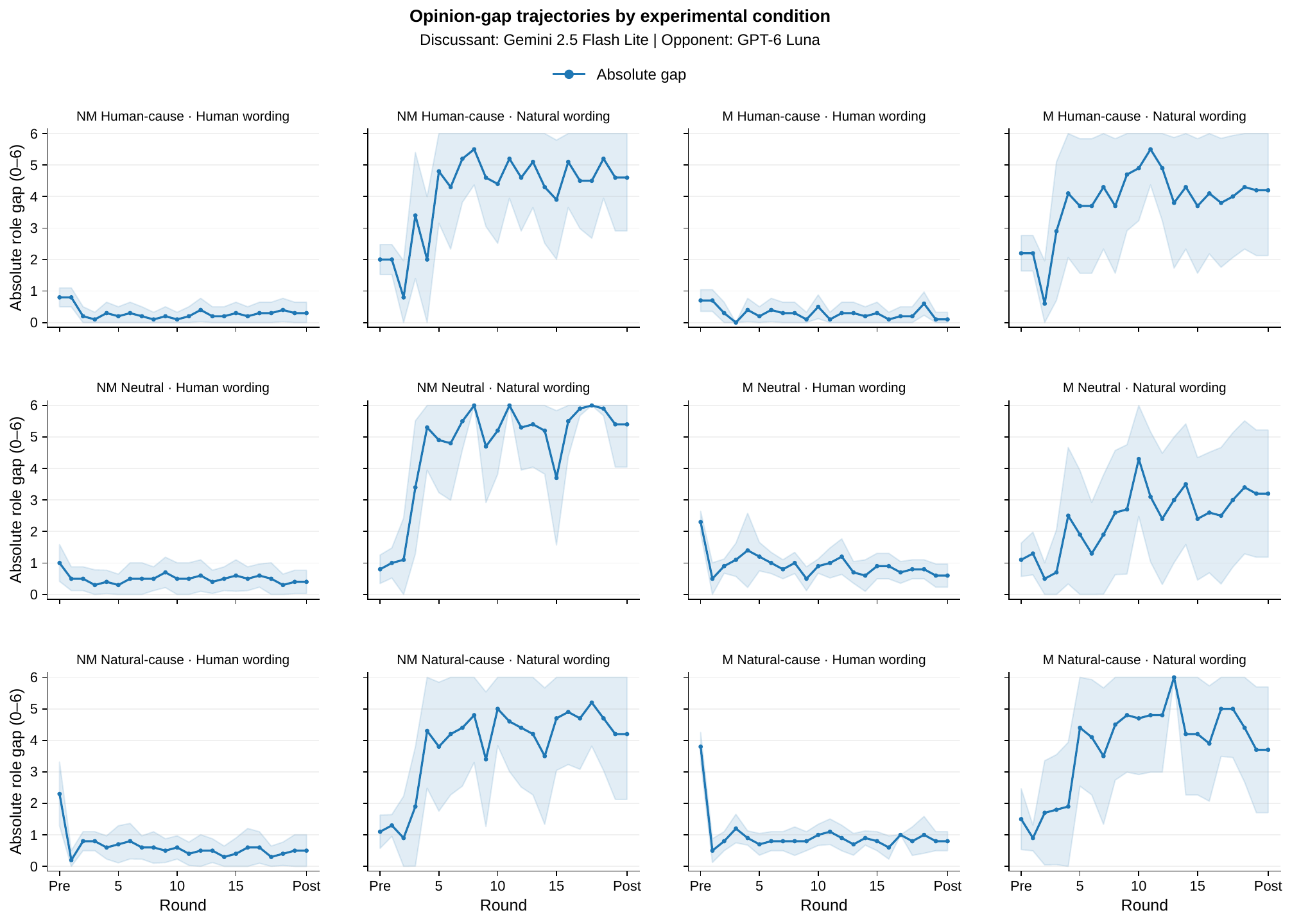}

\caption{Within-pair absolute differences in direct topic ratings over 20 rounds. Ordered pairing: Gemini 2.5 Flash Lite $\to$ GPT-6 Luna.}

\label{fig:app-04-gemini-luna}
\end{figure*}

\begin{figure*}[!tbp]

\centering

\includegraphics[width=0.96\textwidth]{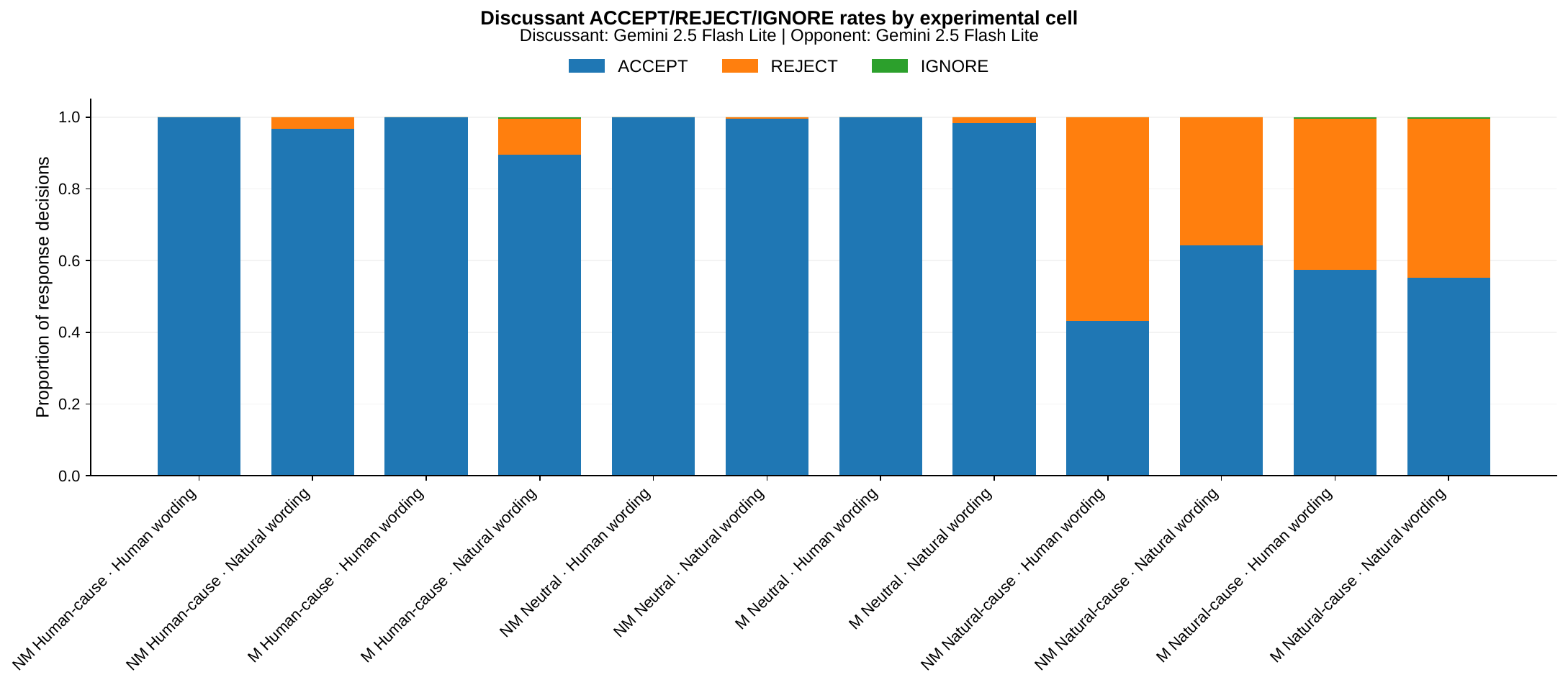}\\[5pt]
\includegraphics[width=0.96\textwidth]{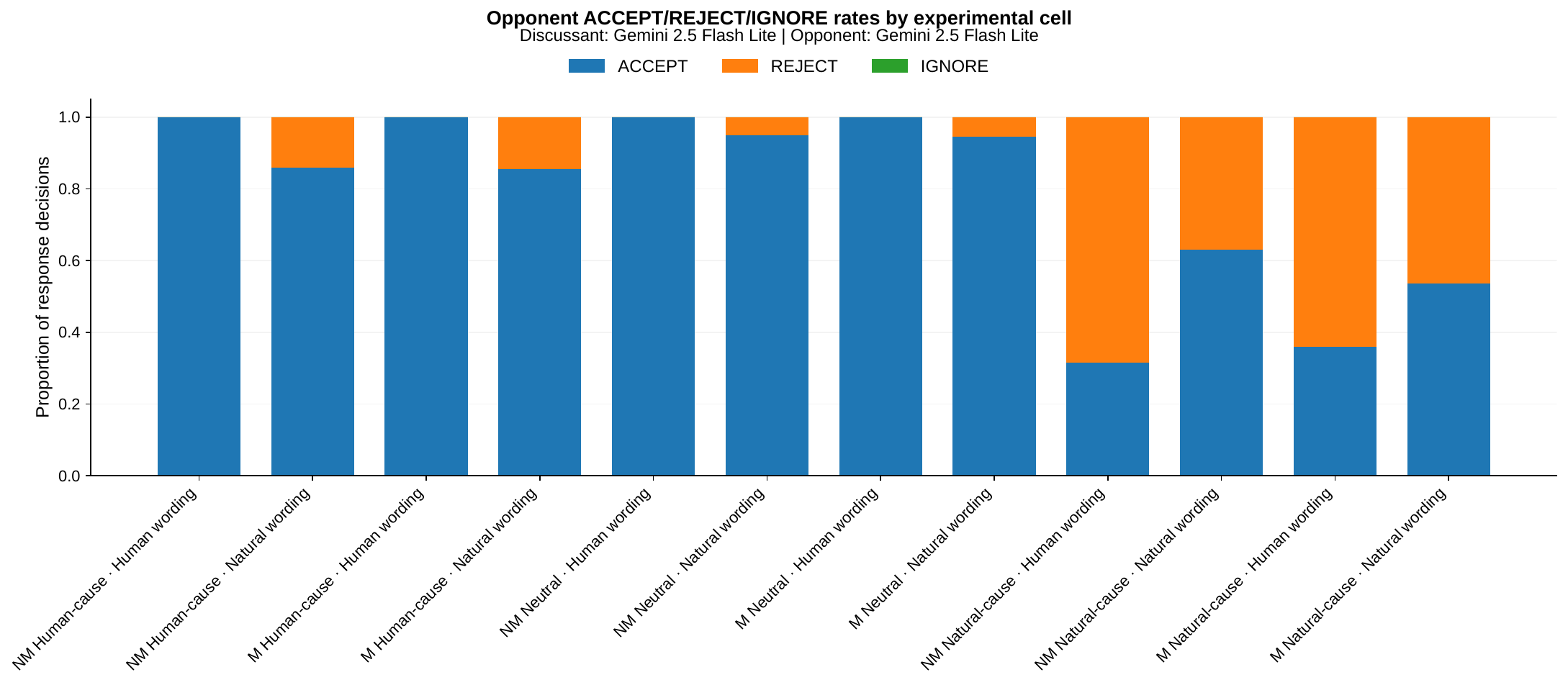}

\caption{ACCEPT/REJECT/IGNORE proportions by exact cell for Discussant (top) and Opponent (bottom). Ordered pairing: Gemini 2.5 Flash Lite $\to$ Gemini 2.5 Flash Lite.}

\label{fig:additional-decisions}

\label{fig:app-05-gemini-gemini}
\end{figure*}

\begin{figure*}[!tbp]

\centering

\includegraphics[width=0.96\textwidth]{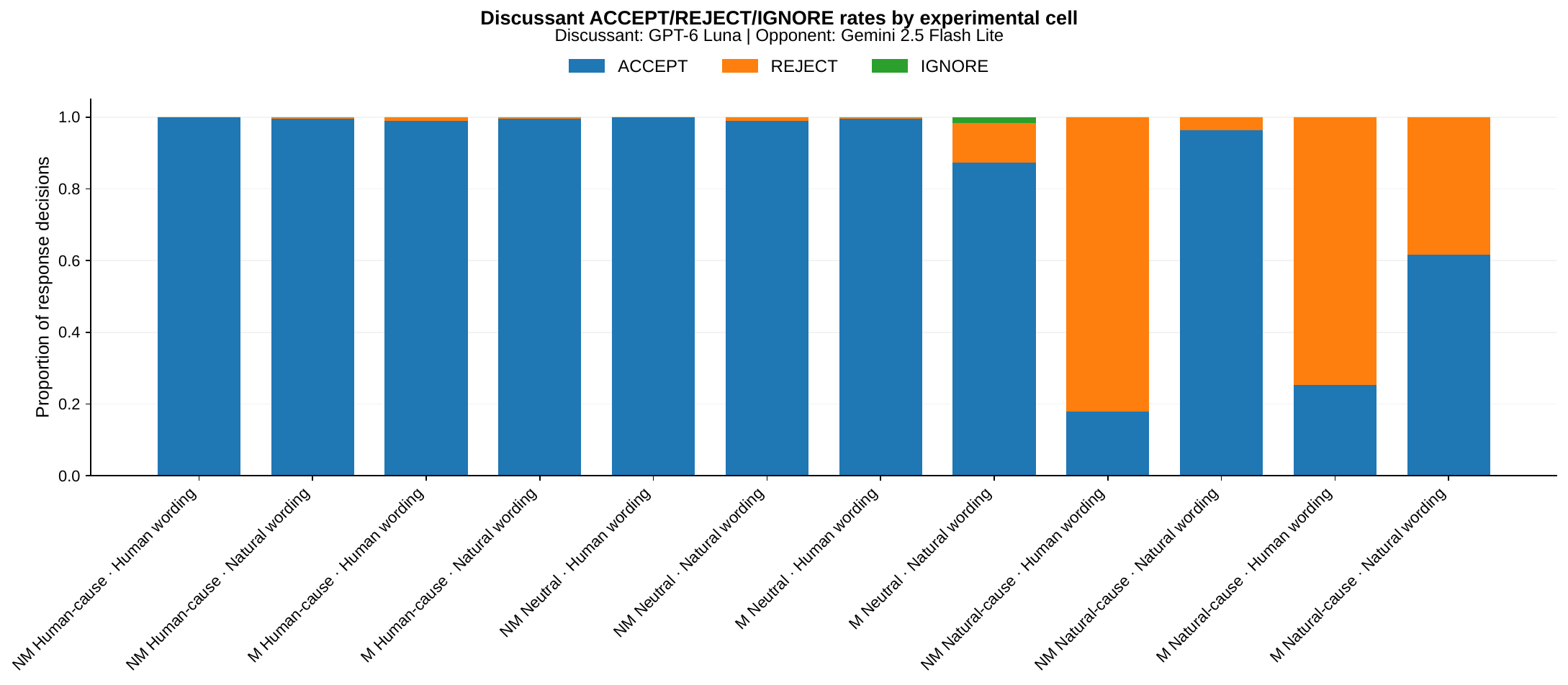}\\[5pt]
\includegraphics[width=0.96\textwidth]{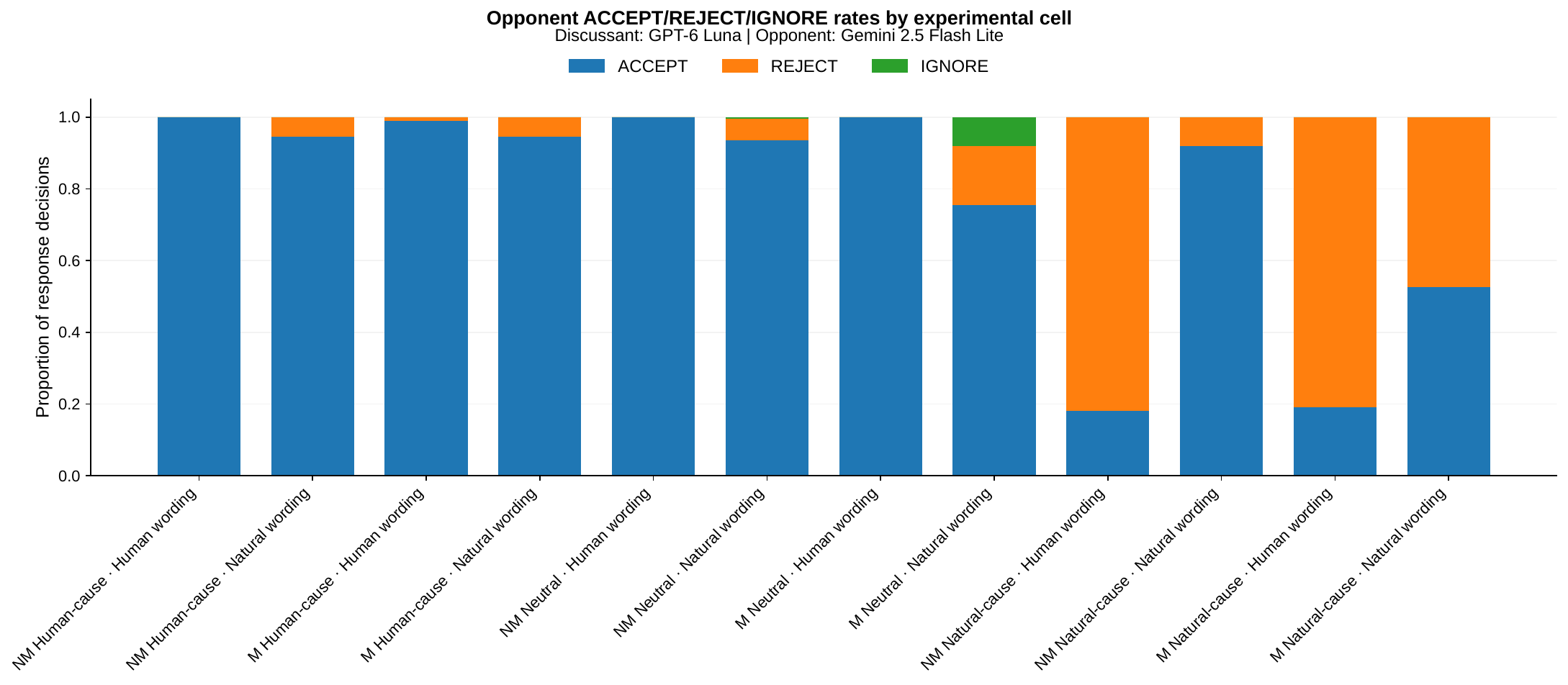}

\caption{ACCEPT/REJECT/IGNORE proportions by exact cell for Discussant (top) and Opponent (bottom). Ordered pairing: GPT-6 Luna $\to$ Gemini 2.5 Flash Lite.}

\label{fig:app-05-luna-gemini}
\end{figure*}

\begin{figure*}[!tbp]

\centering

\includegraphics[width=0.96\textwidth]{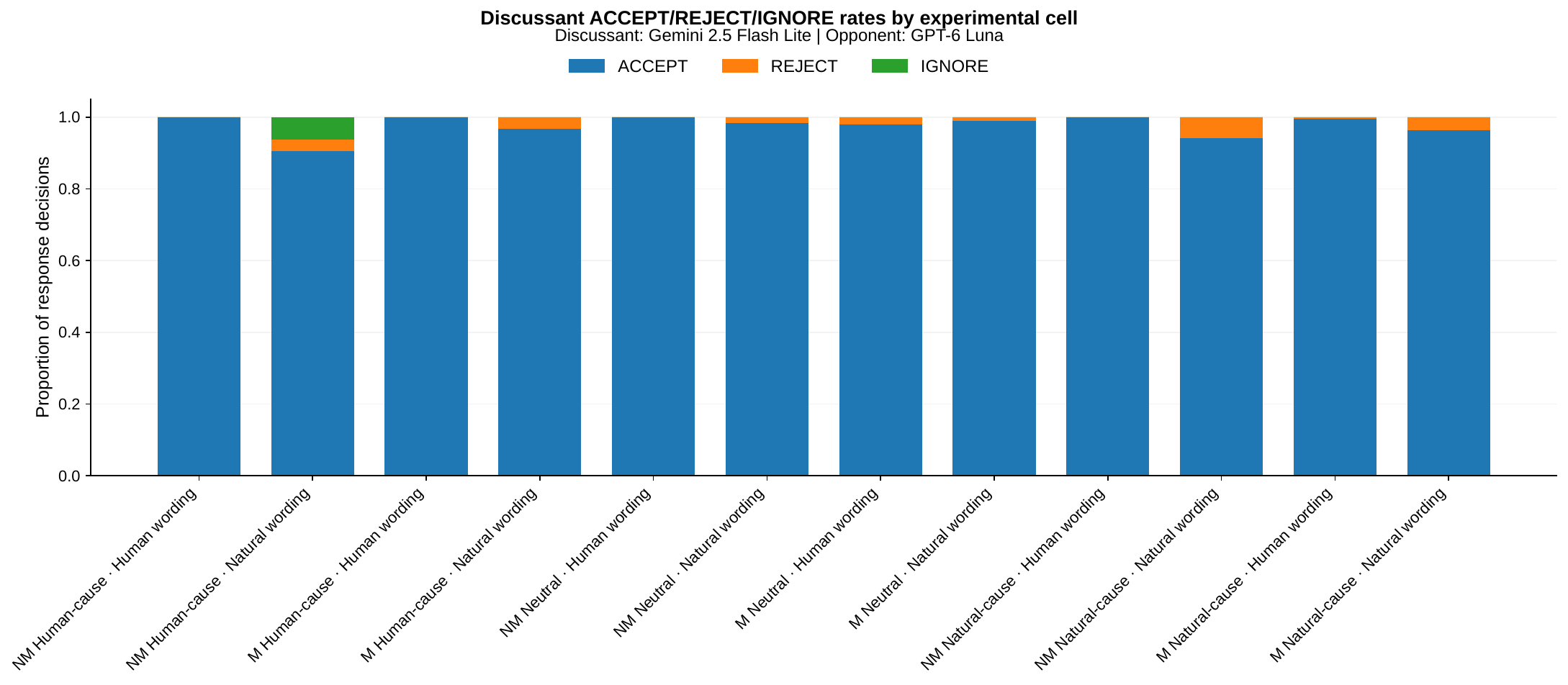}\\[5pt]
\includegraphics[width=0.96\textwidth]{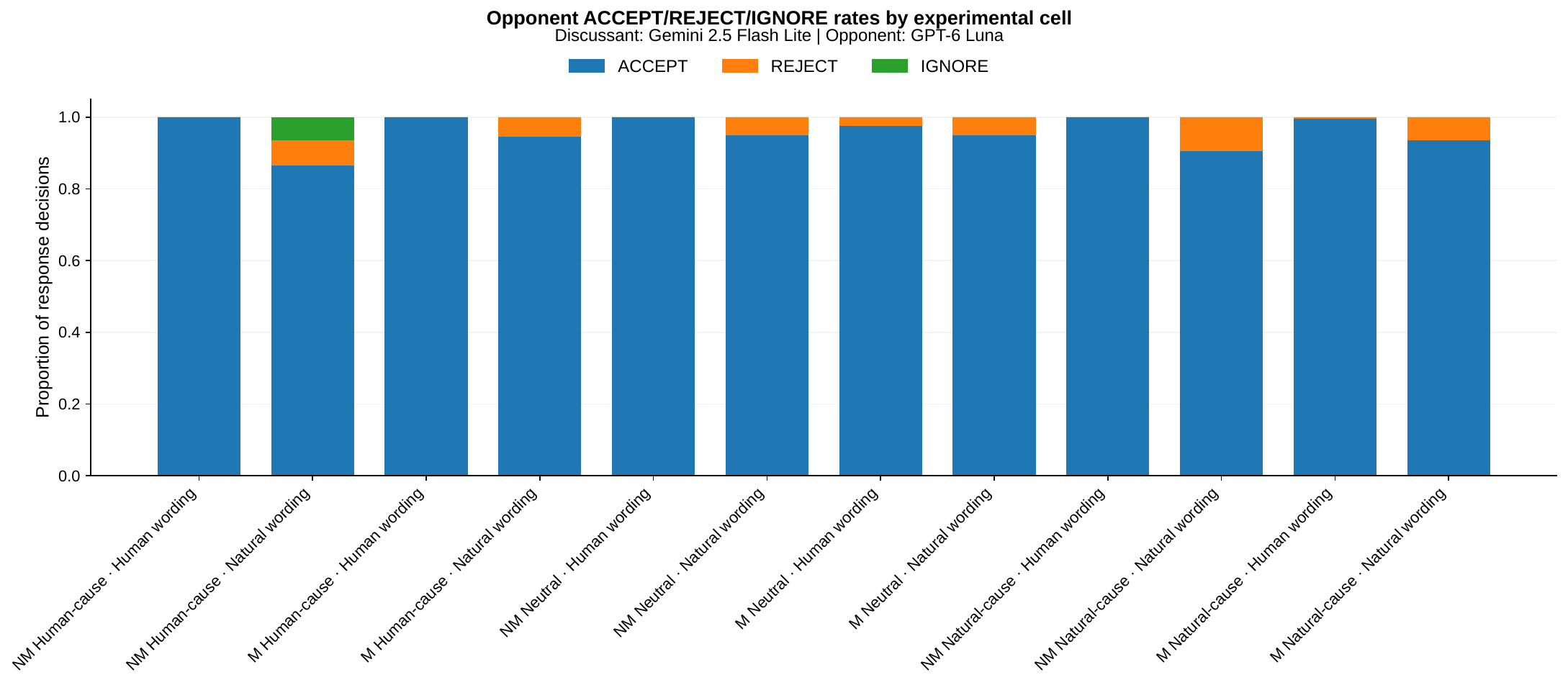}

\caption{ACCEPT/REJECT/IGNORE proportions by exact cell for Discussant (top) and Opponent (bottom). Ordered pairing: Gemini 2.5 Flash Lite $\to$ GPT-6 Luna.}

\label{fig:app-05-gemini-luna}
\end{figure*}

\begin{figure*}[!tbp]

\centering

\includegraphics[width=0.49\textwidth]{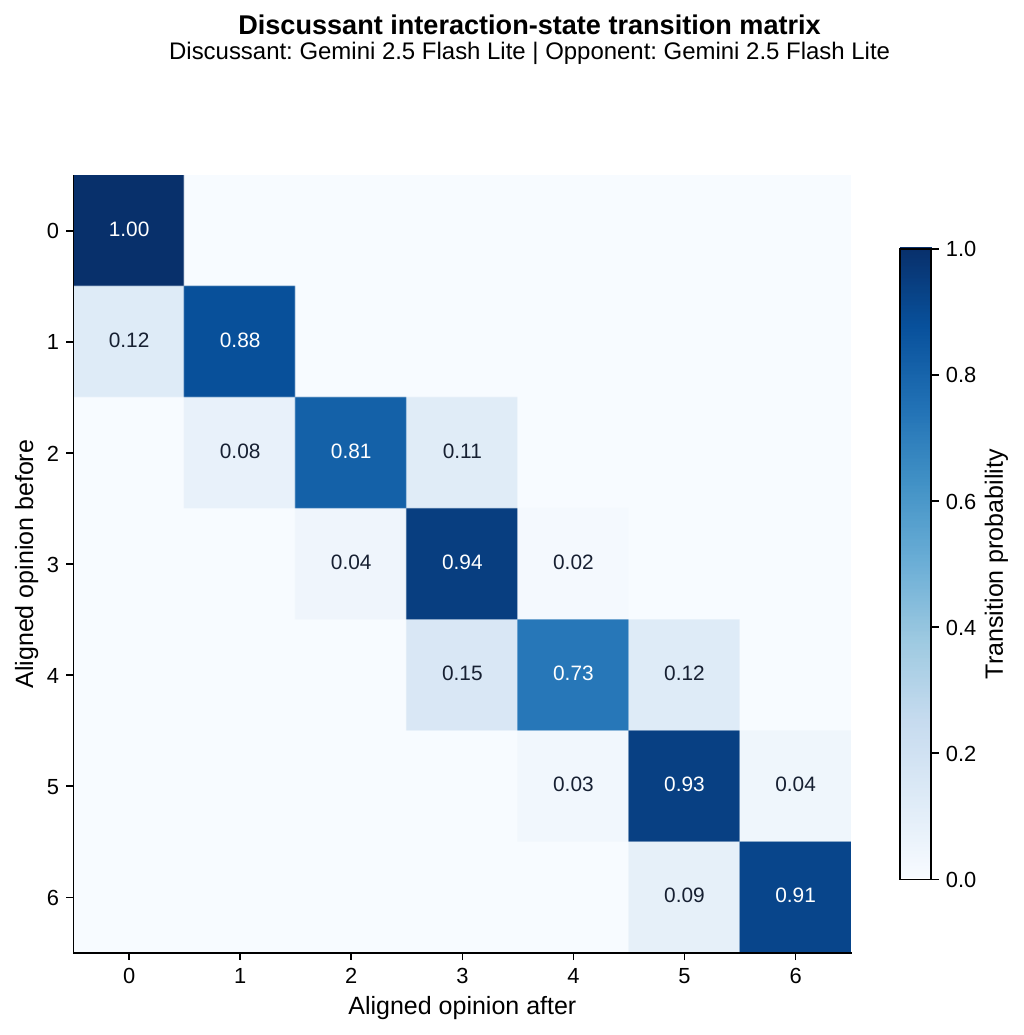}\hfill

\includegraphics[width=0.49\textwidth]{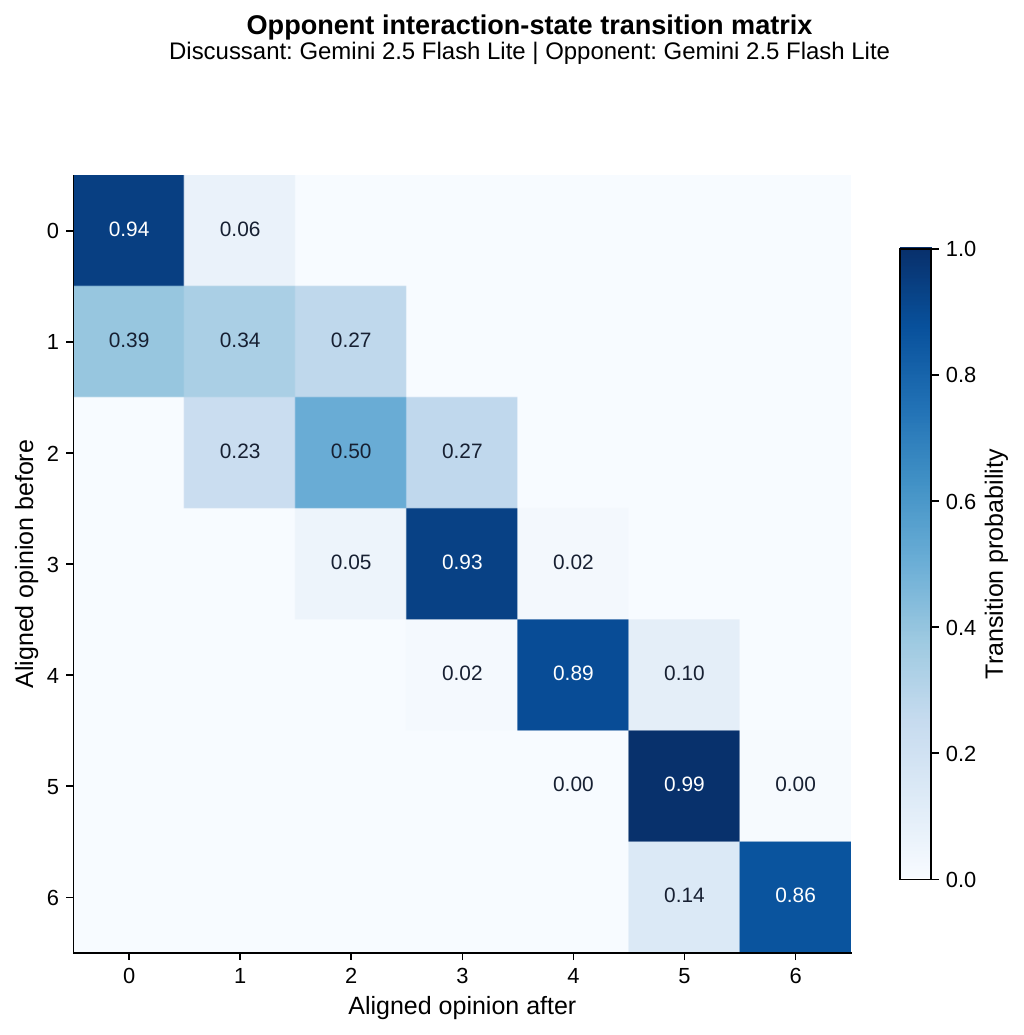}

\caption{Normalized interaction-state transitions for Discussant (top) and Opponent (bottom). Ordered pairing: Gemini 2.5 Flash Lite $\to$ Gemini 2.5 Flash Lite.}

\label{fig:additional-matrices}

\label{fig:app-06-gemini-gemini}
\end{figure*}

\begin{figure*}[!tbp]

\centering

\includegraphics[width=0.49\textwidth]{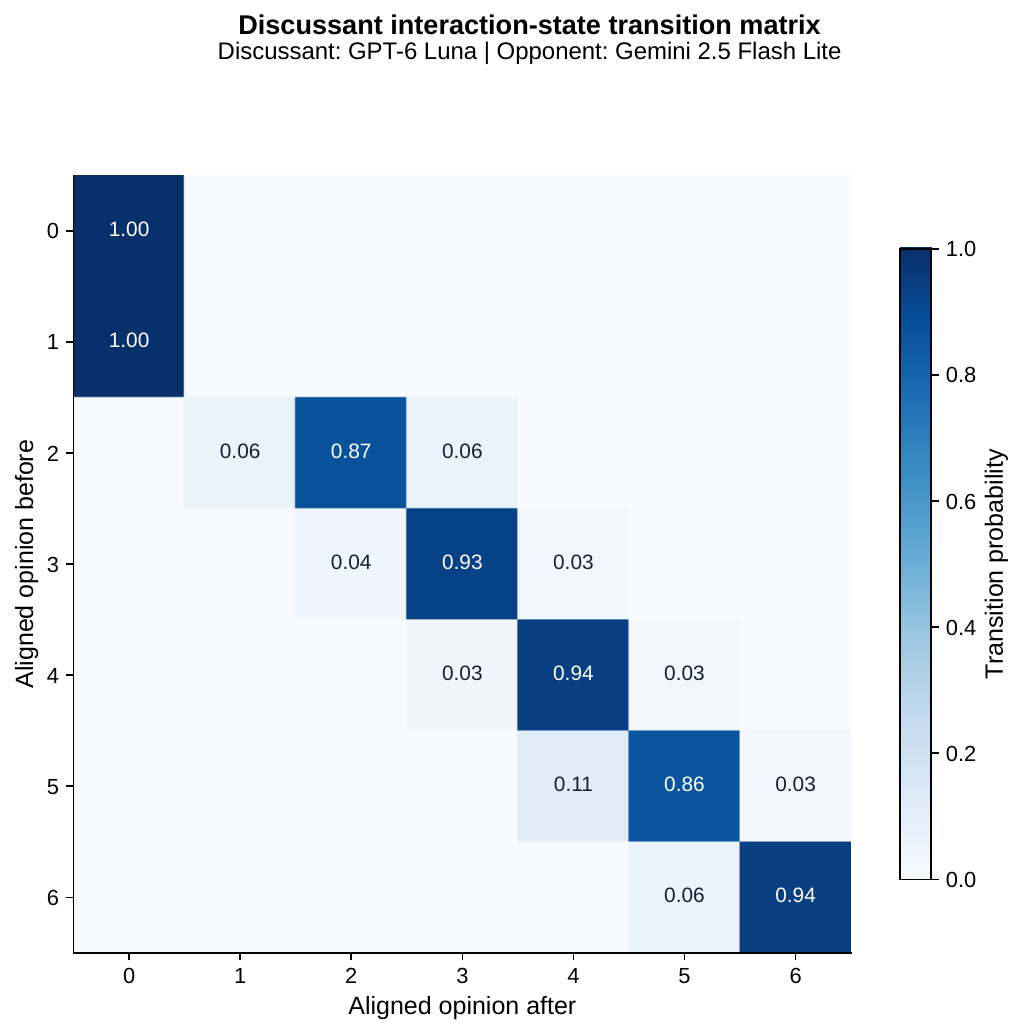}\hfill

\includegraphics[width=0.49\textwidth]{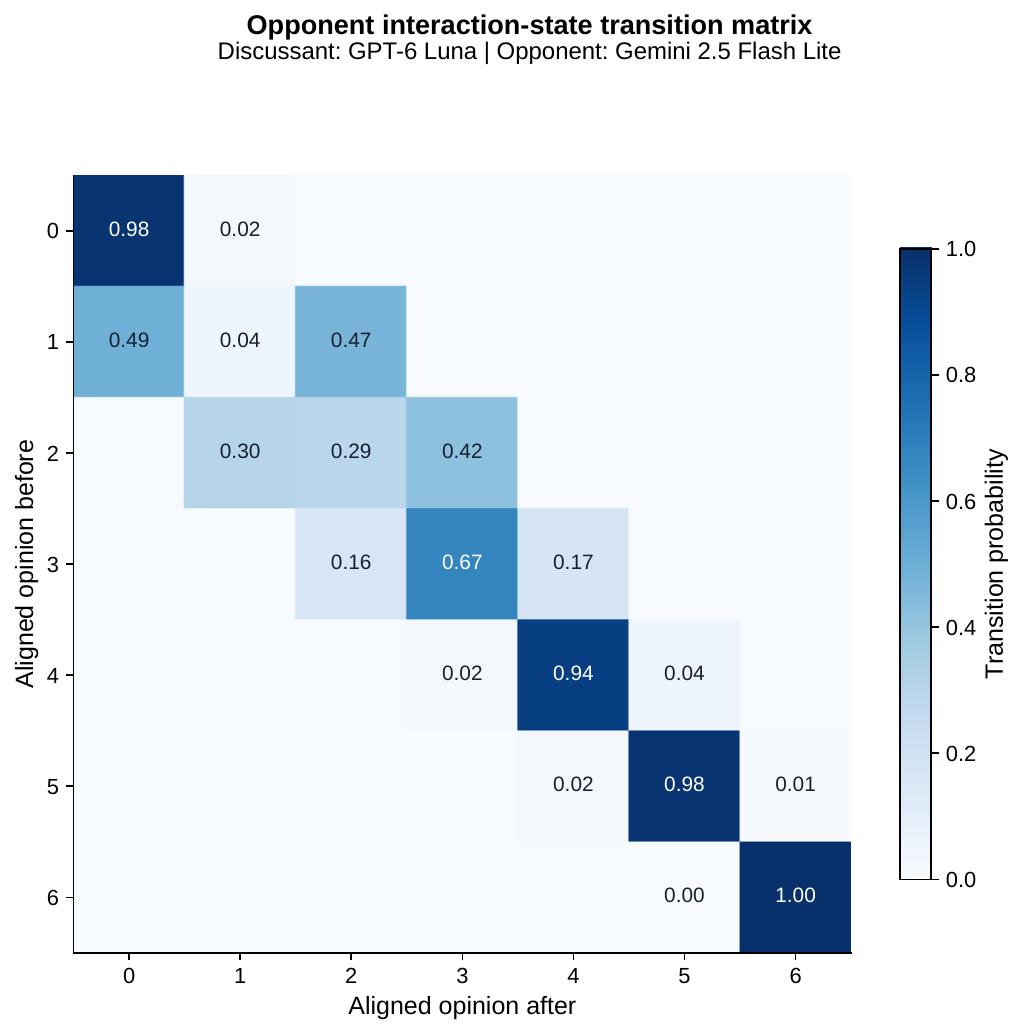}

\caption{Normalized interaction-state transitions for Discussant (top) and Opponent (bottom). Ordered pairing: GPT-6 Luna $\to$ Gemini 2.5 Flash Lite.}

\label{fig:app-06-luna-gemini}
\end{figure*}

\begin{figure*}[!tbp]

\centering

\includegraphics[width=0.49\textwidth]{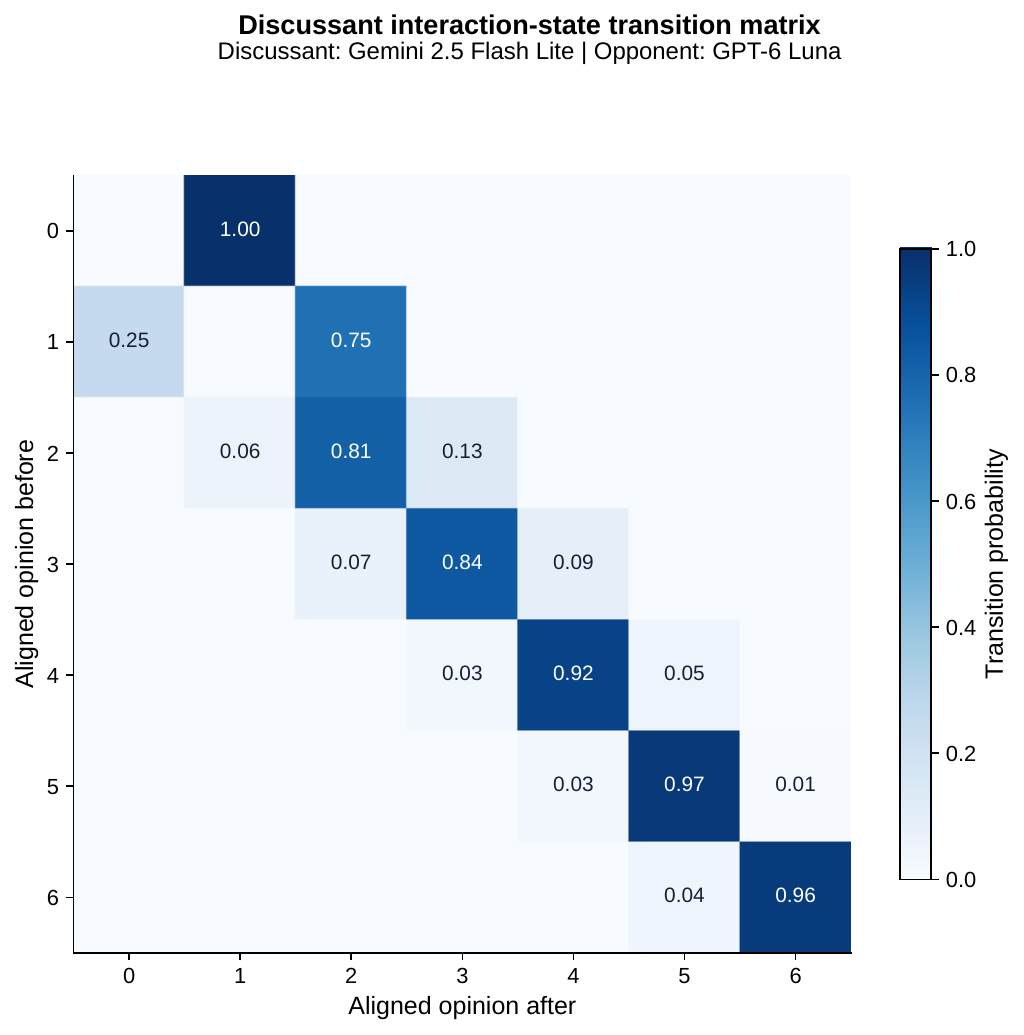}\hfill

\includegraphics[width=0.49\textwidth]{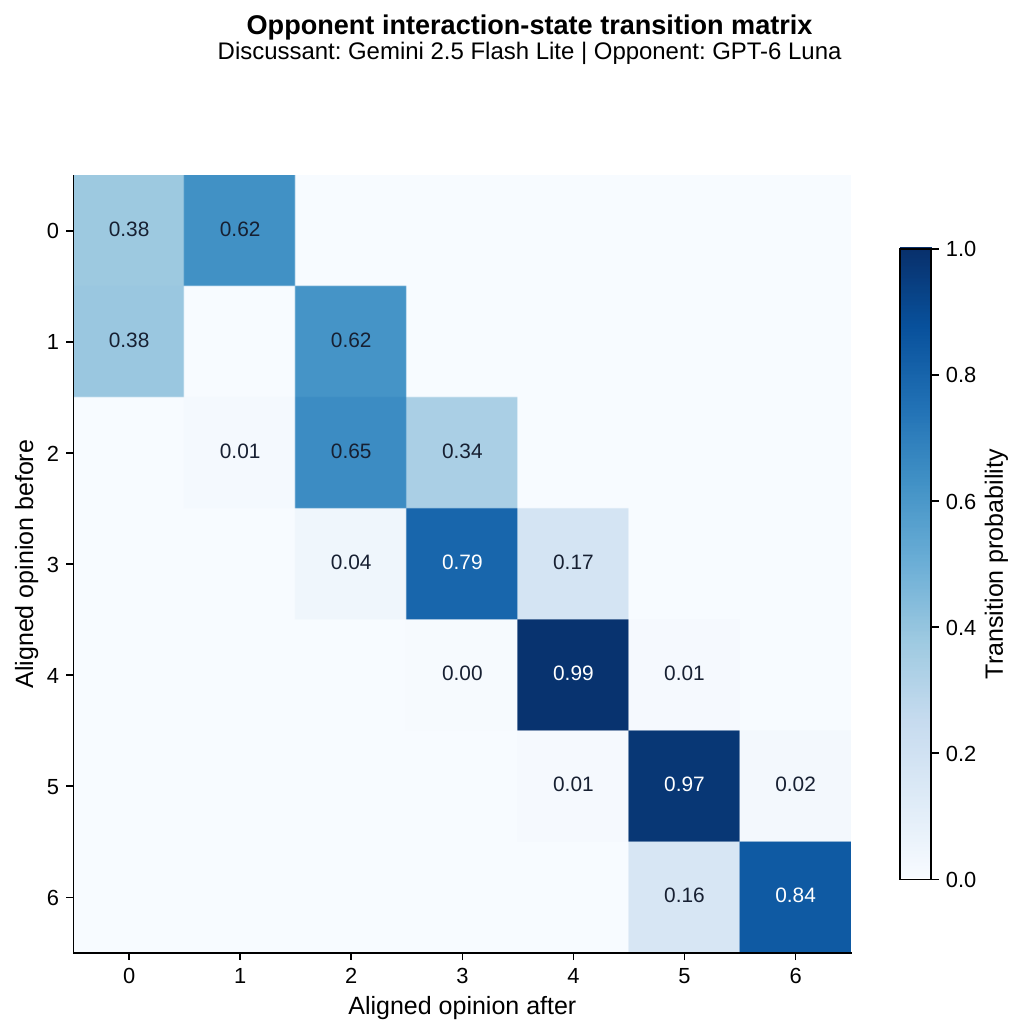}

\caption{Normalized interaction-state transitions for Discussant (top) and Opponent (bottom). Ordered pairing: Gemini 2.5 Flash Lite $\to$ GPT-6 Luna.}

\label{fig:app-06-gemini-luna}
\end{figure*}

\begin{figure*}[!tbp]

\centering

\includegraphics[width=0.49\textwidth]{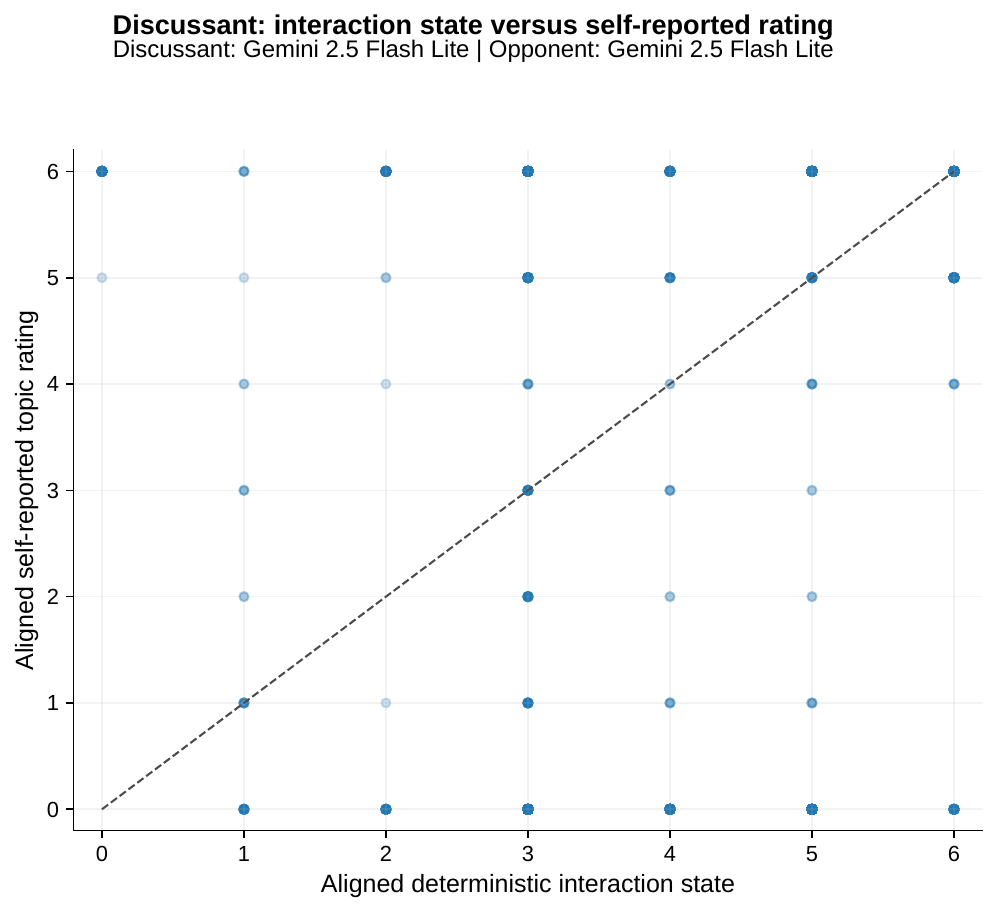}\hfill

\includegraphics[width=0.49\textwidth]{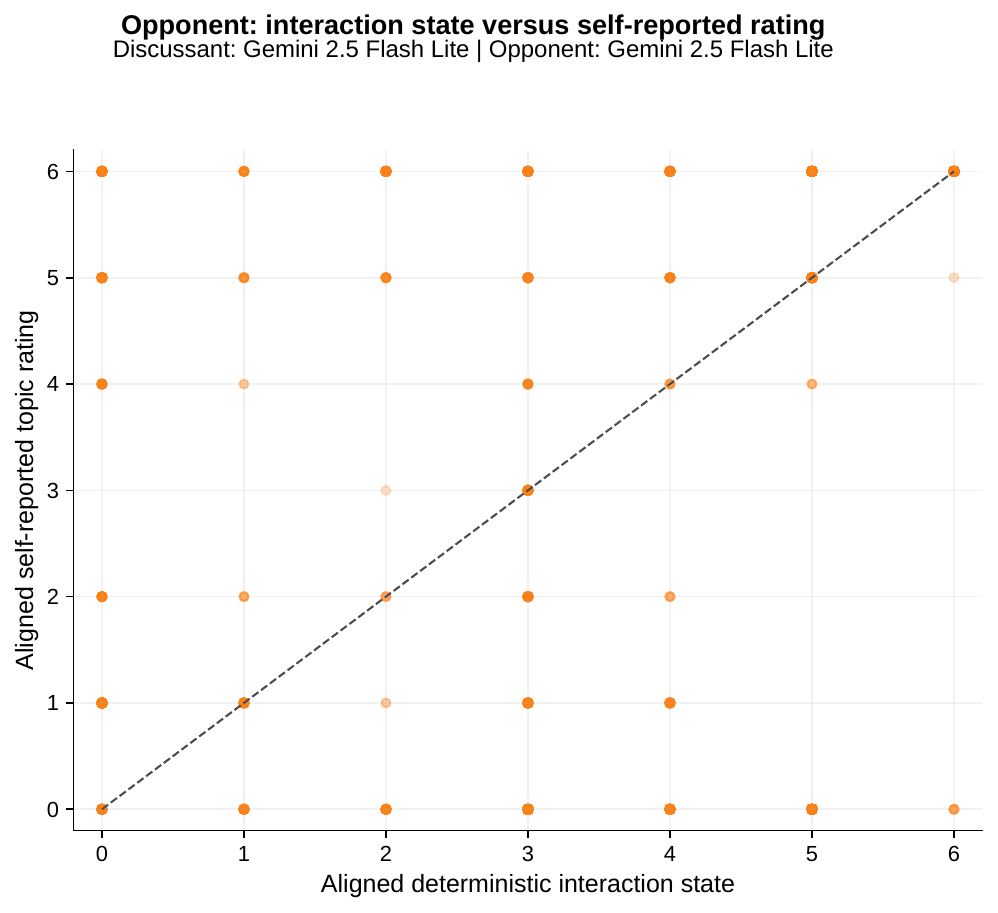}

\caption{Deterministic interaction state versus direct topic rating for Discussant (top) and Opponent (bottom). Ordered pairing: Gemini 2.5 Flash Lite $\to$ Gemini 2.5 Flash Lite.}

\label{fig:additional-scatter}

\label{fig:app-07-gemini-gemini}
\end{figure*}

\begin{figure*}[!tbp]

\centering

\includegraphics[width=0.49\textwidth]{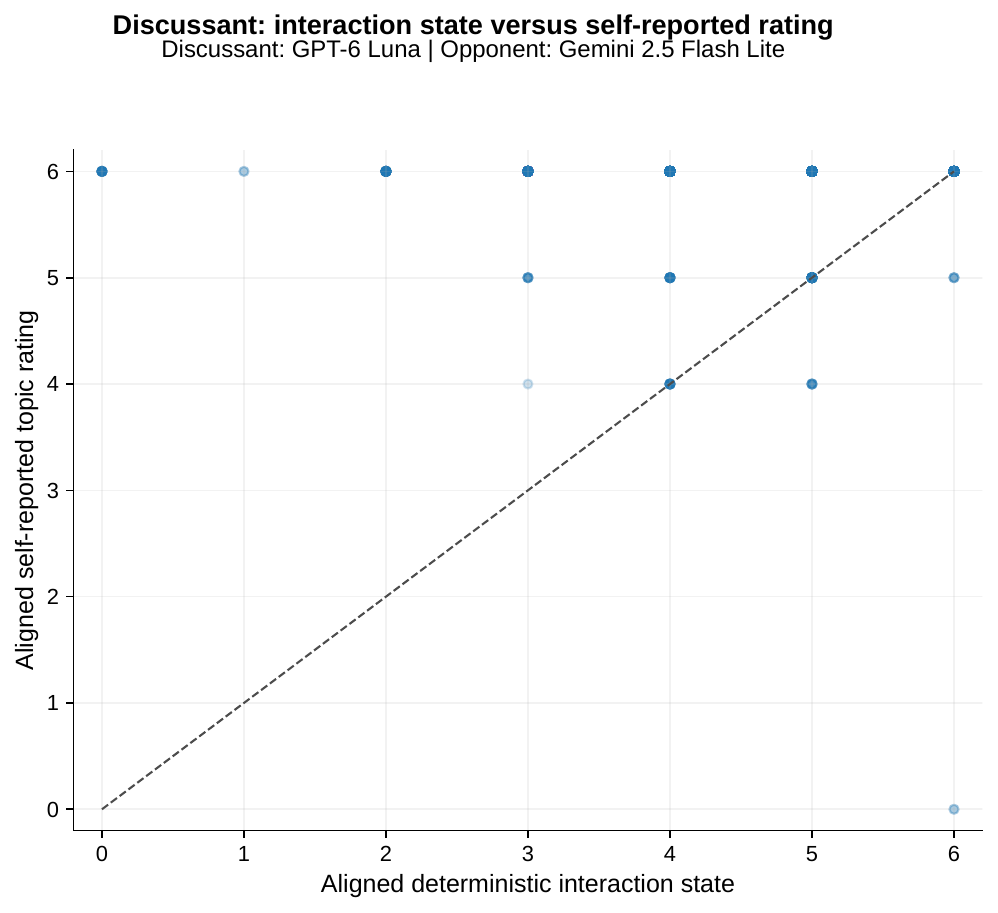}\hfill

\includegraphics[width=0.49\textwidth]{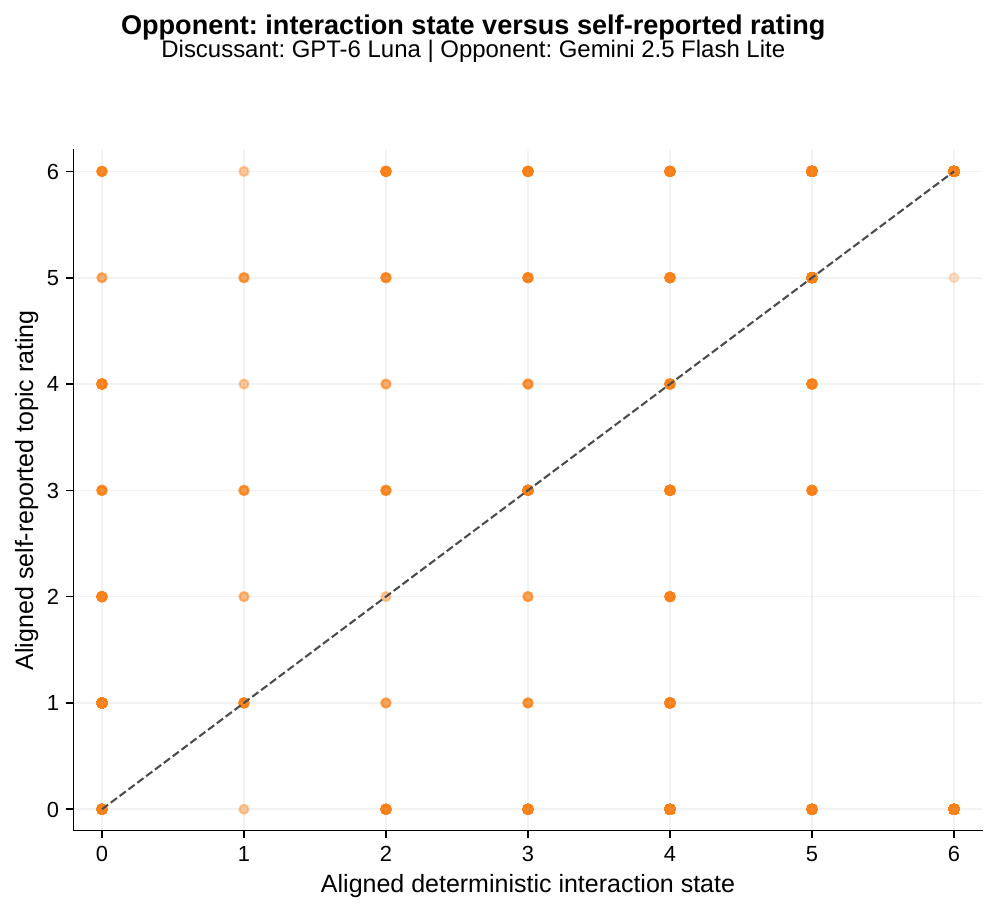}

\caption{Deterministic interaction state versus direct topic rating for Discussant (top) and Opponent (bottom). Ordered pairing: GPT-6 Luna $\to$ Gemini 2.5 Flash Lite.}

\label{fig:app-07-luna-gemini}
\end{figure*}

\begin{figure*}[!tbp]

\centering

\includegraphics[width=0.49\textwidth]{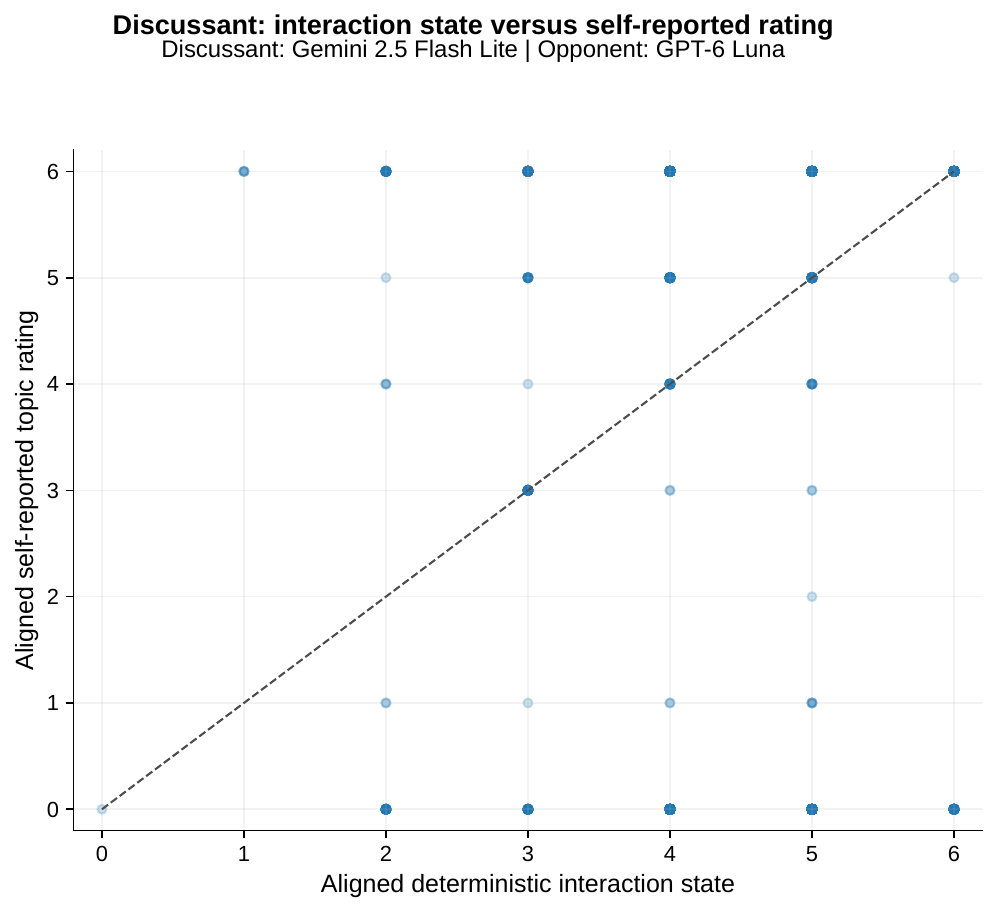}\hfill

\includegraphics[width=0.49\textwidth]{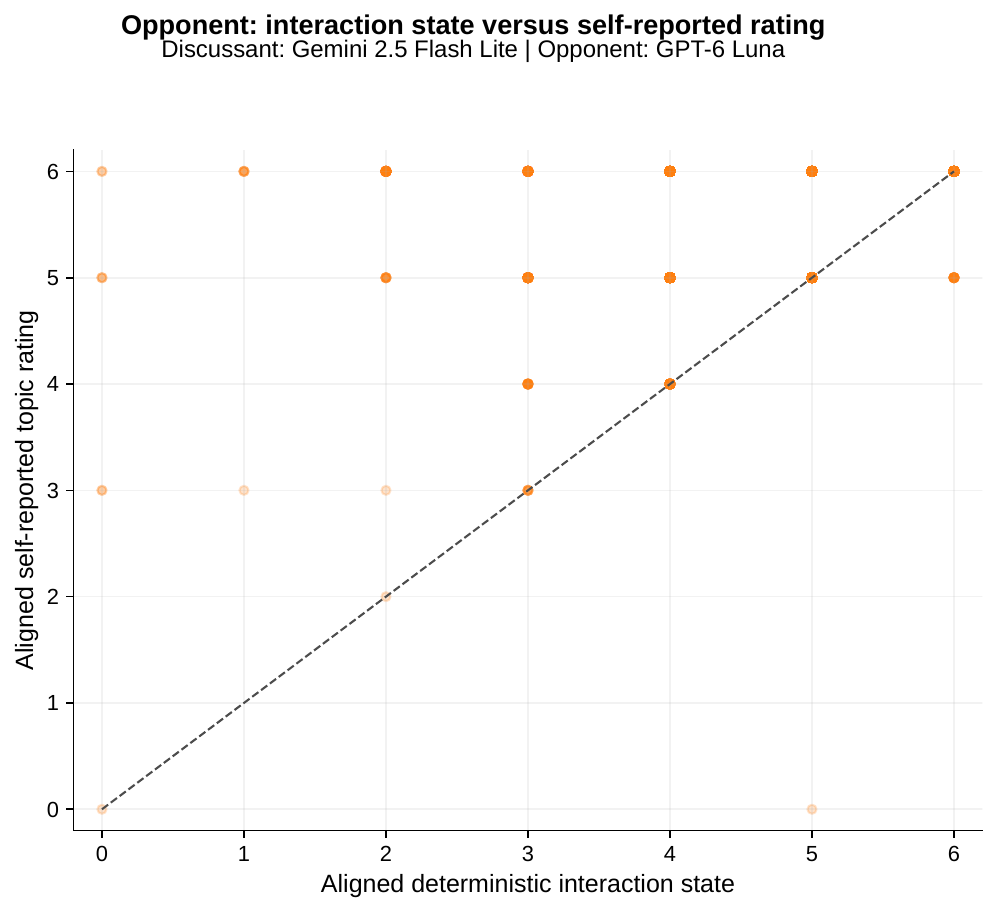}

\caption{Deterministic interaction state versus direct topic rating for Discussant (top) and Opponent (bottom). Ordered pairing: Gemini 2.5 Flash Lite $\to$ GPT-6 Luna.}

\label{fig:app-07-gemini-luna}
\end{figure*}

\begin{figure*}[!tbp]
\centering
\includegraphics[width=0.90\textwidth]{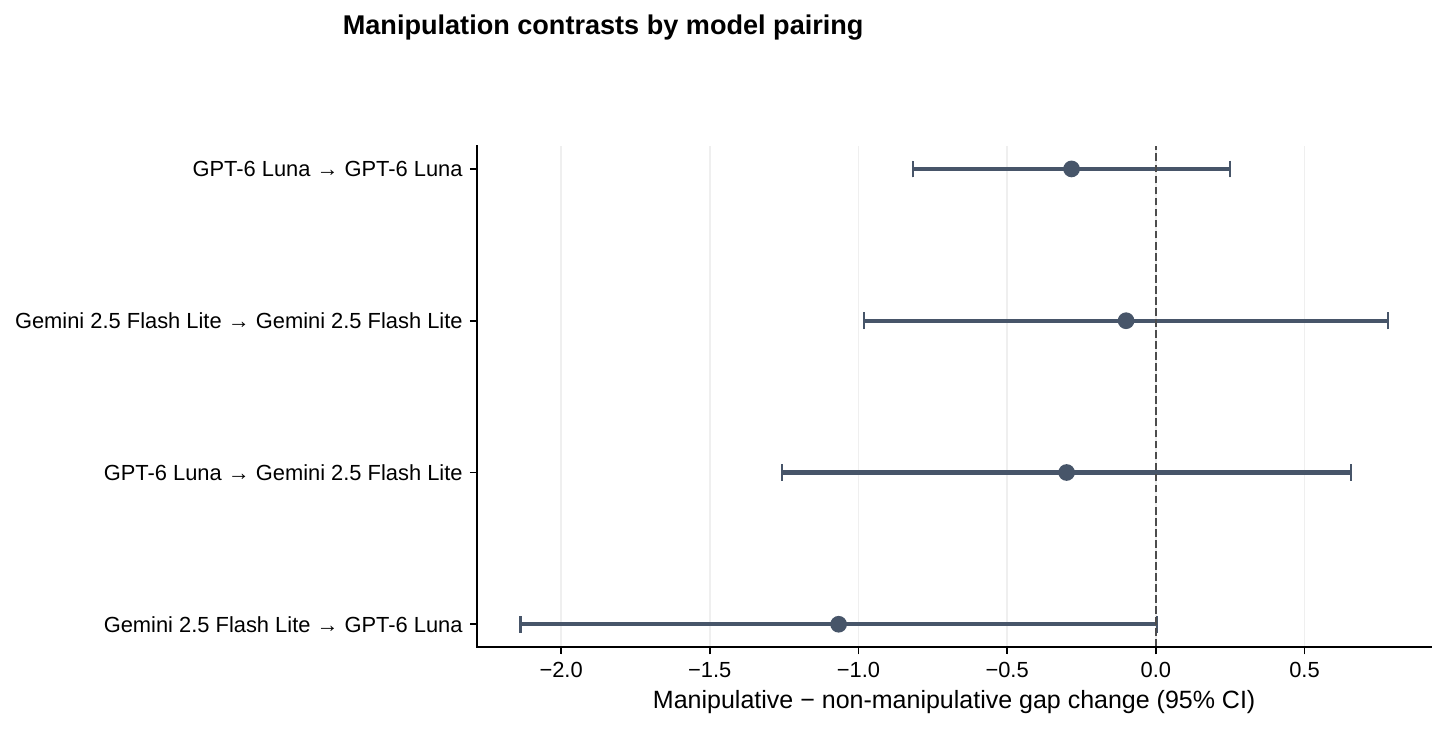}
\caption{Exploratory difference in mean pre/post absolute-gap change between manipulative and non-manipulative Opponent prompts within each ordered model pairing. Points and bars show difference of means and unadjusted Welch 95\% confidence intervals. Negative differences indicate a more convergent (or less divergent) average gap under manipulation, not necessarily persuasion of the Discussant. All four intervals include zero; the pairing-by-manipulation interaction test is not statistically significant.}
\label{fig:cross-manipulation}
\end{figure*}

\end{document}